\documentclass[preprint,12pt]{elsarticle}
\usepackage{amssymb}
\usepackage{amsthm}
\usepackage{amsmath}
\usepackage{amsfonts,amstext,amsmath,amssymb}
\usepackage{hyperref}
\usepackage{multirow}
\usepackage{color}
\usepackage{subfig}
\usepackage{diagbox}
\usepackage[figuresright]{rotating} 
\usepackage{lineno}
\newcommand{\bx}{\boldsymbol{x}}
\newcommand{\bp}{\boldsymbol{p}}
\def\somega#1{\{\omega_#1\}}

\def\somega#1{\{\omega_#1\}}

\begin{document}

\begin{frontmatter}
\title{Application of belief functions to medical image segmentation: A review}

\author[utc,litis]{Ling Huang} \author [litis]{Su Ruan}
\author [utc,iuf]{Thierry Den{\oe}ux}

\address[utc]{Heudiasyc, CNRS, Universit\'e de technologie de Compi\`egne, France}

\address[litis]{Quantif, LITIS, University of Rouen Normandy, France}

\address[iuf]{Institut universitaire de France, France}

\begin{abstract}
The investigation of uncertainty is of major importance in risk-critical applications, such as medical image segmentation. Belief function theory, a formal framework for uncertainty analysis and multiple evidence fusion, has made significant contributions to medical image segmentation, especially since the development of deep learning. In this paper, we provide an introduction to the topic of medical image segmentation methods using belief function theory. We classify the methods according to the fusion step and explain how information with uncertainty or imprecision is modeled and fused with belief function theory. In addition, we discuss the challenges and limitations of present belief function-based medical image segmentation and propose orientations for future research. Future research could investigate both belief function theory and deep learning to achieve more promising and reliable segmentation results. 
\end{abstract}


\end{frontmatter}

\section{Introduction}

The precise delineation of the target lesion on the medical image is essential for optimizing disease treatment. In clinical routine, this segmentation is performed manually by physicians and has shortcomings. First, the pixel or voxel-level segmentation process is time-consuming, especially with 3D images. Second, the segmentation performances are limited by the quality of medical images, the  difficulty of the disease, and the domain knowledge of the experts. Thus, physicians always diagnose a disease by summarizing multiple sources of information. The advances in medical imaging machines and technology now allow us to obtain medical images in several modalities, such as Magnetic Resonance Imaging (MRI)/Positron Emission Tomography(PET), multi-sequence MRI, or PET/ Computed Tomography (CT). The different modalities provide different information about cancer and other abnormalities in the human body. Combining these modalities makes it possible to segment the organ and lesion region in a reliable way.

Classical medical image segmentation
approaches~\cite{batenburg2009adaptive,kimmel2003fast,onoma2014segmentation,salvador2004determining,suykens1999least,StroblBias,kleinbaum1996logistic}
focus on low-level feature analysis, e.g.,~gray and textual features or
hand-crafted features. Those approaches have limitations in terms of  segmentation accuracy, efficiency, and reliability, which creates a big gap between experimental performance and clinical application. More recently, the success of deep learning in the medical domain brought a lot of contributions to medical image segmentation
tasks~\cite{ronnebergerconvolutional,myronenko20183d,isensee2018nnu,bahdanau2014neural,vaswani2017attention,carion2020end,han2021transformer}.
It first solves the segmentation efficiency problem, allowing for large-scale medical image segmentation, which contributes greatly to its accuracy. 

\begin{figure}
 \centering
\includegraphics[scale=0.6]{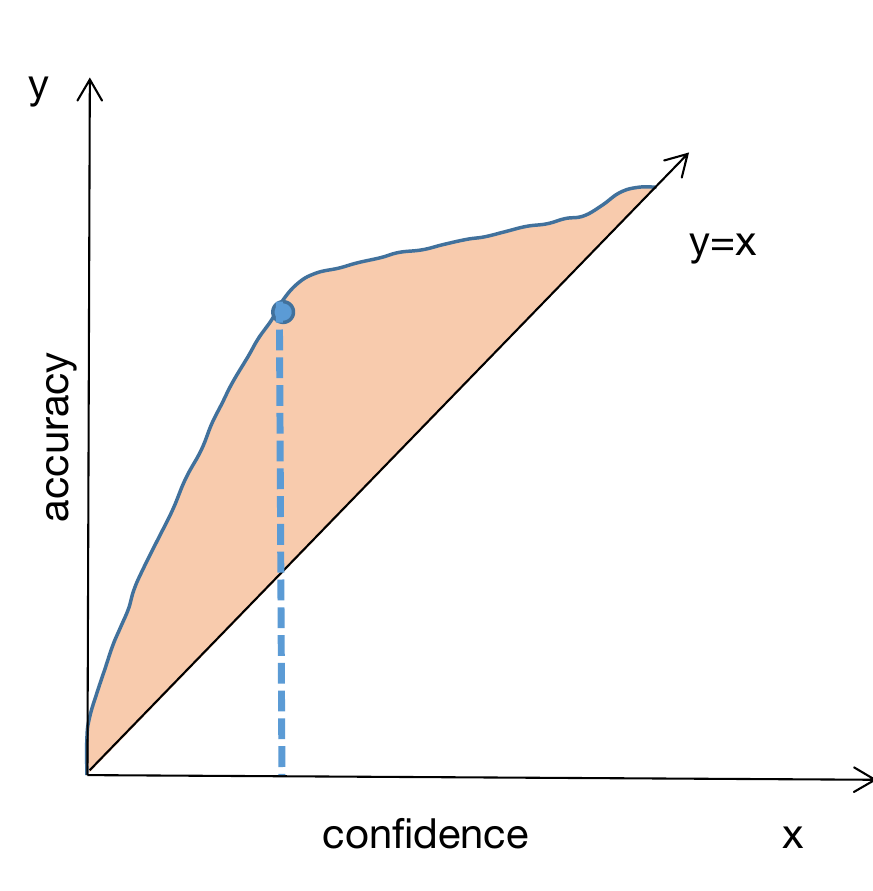}
\caption{Example of a reliability diagram of an overconfidence model.\label{fig:over-confidence}}
\end{figure}

Despite the excellent performance of deep learning-based medical image segmentation methods, doubts about the reliability of the segmentation results still remain~\cite{hullermeier2021aleatoric}, which explains why their application to therapeutic decision-making for complex oncological cases is still  limited. A reliable segmentation model should be well calibrated, i.e., its confidence should match its accuracy. Figure \ref{fig:over-confidence} shows an example of an over-confidence segmentation model that outputs a result with high segmentation accuracy and low confidence. Therefore, a trustworthy representation of uncertainty is desirable and should be considered a key feature of any deep learning method, especially in safety-critical application domains, e.g., medical image segmentation. In general, deep models have two sources of uncertainty: aleatory uncertainty and epistemic uncertainty~\cite{hora1996aleatory,der2009aleatory}. Aleatory uncertainty refers to the notion of randomness, i.e., the variability in an experiment's outcome due to inherently random effects. In contrast, epistemic uncertainty refers to uncertainty caused by a lack of knowledge (ignorance) about the best model, i.e., the ignorance of the learning algorithm or decision-maker. As opposed to uncertainty caused by randomness, uncertainty caused by ignorance can be reduced based on additional information or the design of a suitable learning algorithm.

In medical image segmentation, the uncertainty can be decomposed into three levels~\cite{lakshminarayanan2017simple}. \emph{Pixel/voxel-level} uncertainty is useful for interaction with physicians by providing additional guidance for correction segmentation results. \emph{Instance-level} is the uncertainty aggregate by pixel/voxel-level uncertainty, which can be used to reduce the false discovery rate. \emph{Subject-level} uncertainty offers information on whether the segmentation model is reliable. Early approaches to quantify the segmentation uncertainty quantification were based on Bayesian theory~\cite{hinton1993keeping,mackay1992practical}. The popularity of deep segmentation models has revived  research on model uncertainty estimation and has given rise to specific methods such as variational dropout~\cite{gal2016dropout,tran2018bayesian}, and model ensembles~\cite{lakshminarayanan2017simple,rupprecht2017learning}. However, probabilistic segmentation models capture knowledge in terms of a single probability distribution and cannot distinguish between aleatory and epistemic uncertainty, which limits the exploitation of the results. In this paper, we review a different approach to uncertainty quantification based on belief function theory (BFT)~\cite{dempster1967upper,shafer1976mathematical,denoeux20b} with a specific focus on medical image segmentation. BFT can model epistemic uncertainty directly, which makes it possible to explore the possibility of improving the model's reliability based on the existing uncertainty. Researchers from the medical image segmentation community have been actively involved in the research on BFT for handling uncertain information modeling and
fusion, an approach that has been shown to be very promising for segmenting imperfect medical images~\cite{lian2018joint,ghesu2021quantifying,huang2021belief}.

The rest of the paper will provide an overview of the BFT-based medical image
segmentation methods. Recent review work on medical image segmentation
focus on deep learning-based
methods~\cite{razzak2018deep,hesamian2019deep,liu2021review}, with emphasis on deep feature extraction~\cite{siddique2021u}, multimodal information fusion~\cite{zhou2019review}, embracing imperfect dataset~\cite{ramesh2021review}, etc. This paper will focus on discussing: 
\begin{itemize}
    \item [(1)] how imperfect medical images can be modeled by assigning them the degree of belief and uncertainty directly;
    \item [(2)] how multiple sources of evidence (with conflict) can be fused by BFT;
    \item [(3)] how the above advantages can be merged with the popular deep learning methods to get an accurate and reliable deep segmentation model.
\end{itemize}
To the best of our knowledge, this is the first review that summarizes the existing medical image segmentation methods with uncertainty quantification and multiple evidence fusion. We hope this review will raise the community's awareness of the existing solutions for imperfect medical image data segmentation and further inspire researchers to explore the possibility of exploiting the benefits of both BFT and deep learning to make automatic segmentation methods reliable and interpretable. 

We organize this paper as follows: Section \ref{sec:medical_image_segmentation} summarizes the development of existing medical image segmentation methods and their limitations. Section \ref{sec:BFT} introduces the fundamentals of BFT. Section \ref{sec:mass_finction} introduces the BFT-based methods to model uncertainty, with Sections \ref{subsec:supervised} and \ref{subsec:unsupervised} present supervised and unsupervised methods to model uncertainty, respectively. Section \ref{sec:BFT-based fusion} gives an overview of BFT-based medical image segmentation methods with Sections \ref{subsec: single-clasisifer} and \ref{subsec: multi-clasisifer} present the BFT-based medical image segmentation with, respectively, single and multiple classifiers or clustering algorithms. Section \ref{sec:conclusion} concludes this review and gives some potential research  directions.

\textbf{Search criterion}. To identify related contributions, we mainly retrieved papers containing ``medical image segmentation'' and ``belief function theory'' or ``Dempster--Shafer theory'' or ``evidence theory'' in the title or abstract from IEEE, Springer, PubMed, Google Scholar, and ScienceDirect. Additionally, conference proceedings for NIPS, CVPR, ECCV, ICCV, MICCAI, and ISBI were searched based on the titles of papers. Papers that do not primarily focus on medical image segmentation problems were excluded.

 \section{Medical image segmentation}
\label{sec:medical_image_segmentation}

Medical image segmentation involves the extraction of regions of interest (ROI) from 2D/3D image data (e.g.,~pathological or optical imaging with color mages, MRI, PET, and CT scans. The main goal of segmenting medical images is to identify areas of cancer and other abnormalities in the human body, for example, brain tissue and tumor segmentation, the interior of the human body (such as lung, spinal canal, and vertebrae), skin and cell lesion segmentation with optical imaging with color. Previously, segmenting medical images was time-consuming for physicians, while recent advances in machine learning techniques, especially deep learning, make it easier to perform routine tasks.

\subsection{Traditional approaches to medical images segmentation}

Early image segmentation methods used the information provided by the image itself, e.g.,~gray, textual, contrast and histogram features, and segmenting ROI-based threshold~\cite{batenburg2009adaptive}, edge detection~\cite{kimmel2003fast}, graph partitioning~\cite{onoma2014segmentation}, and clustering~\cite{salvador2004determining}. More recently, researchers have been interested in hard-crafted features, e.g.,~Scale Invariant Feature Transform (SIFT)~\cite{lowe1999object}, Features from Accelerated Segment Test (FAST)~\cite{rosten2006machine} and Geometric hashing~\cite{mian2006three}, and segmenting ROI using machine learning-based methods such as support vector machine (SVM)~\cite{suykens1999least}, random forest (RF)~\cite{StroblBias} and logistic regression (LR)~\cite{kleinbaum1996logistic}. Those methods have attracted great interest for a while, but their accuracy cannot meet clinical application requirements.

\subsection{Deep learning-based methods}

Long et al.~\cite{long2015fully} were the first authors to show that a fully convolutional network (FCN) could be trained end-to-end for semantic segmentation, exceeding the state-of-the-art when the paper was published in 2015. UNet~\cite{ronnebergerconvolutional}, a successful modification and extension of FCN, has become the most popular model for medical image segmentation in recent years. Based on UNet, research for deep learning-based medical image segmentation can be summarized in two major directions: feature representation and model optimization.

To better represent input information, well-designed DNNs with encoder-decoder architecture (e.g.,~3D UNet~\cite{cciccek20163d},
Res-UNet~\cite{xiao2018weighted}, Dense-UNet~\cite{guan2019fully},
MultiResUNet~\cite{ibtehaz2020multiresunet}, Deep3D-UNet~\cite{zhu2018anatomynet}, V-Net~\cite{milletari2016v}, nnUNet~\cite{isensee2018nnu}) have been proposed and have achieved good performance. Recently,  popular ideas such as, e.g.,~residual connection~\cite{he2016deep},
attention mechanism~\cite{bahdanau2014neural} and transformer
mechanism~\cite{vaswani2017attention,carion2020end,han2021transformer}, have
achieved promising performance with DNNs, which also led to the contributions
in medical image domain, e.g.,~Residual-UNet~\cite{ronnebergerconvolutional,xiao2018weighted}, Attention
UNet~\cite{oktay2018attention,trebing2021smaat} and
Transformer UNet~\cite{cao2021swin,hatamizadeh2022unetr,hatamizadeh2022swin}. 

As for the optimization of the DNN-based medical image segmentation models, researchers in the medical image segmentation domain tend to use the Dice Loss or a combination of Dice loss and cross-entropy as a total loss~\cite{milletari2016v} instead of the classical loss functions (e.g.,~Cross-Entropy Loss, Mean Square Loss) to handle the unbalanced label distribution problem, with more focuses on mining the foreground area. Based on the Dice loss, some variants, such as the  Generalized Dice Loss~\cite{sudre2017generalised}, Tversky's Loss~\cite{salehi2017tversky} and Contrastive Loss~\cite{chen2020simple}, have been proposed to further solve the unbalanced label problem.

\begin{figure}
\centering
\includegraphics[width=0.5\textwidth]{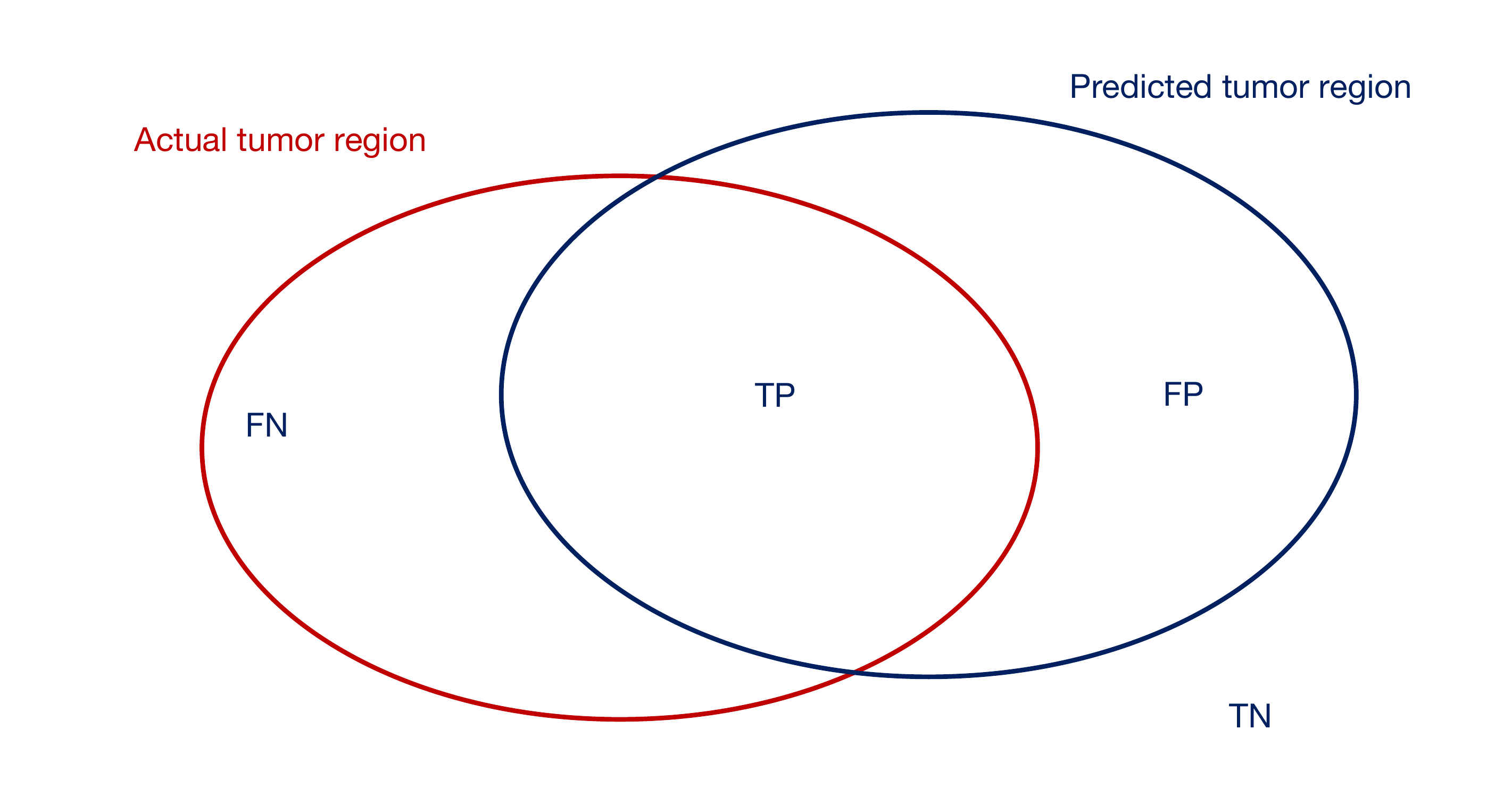}
\caption{Interpretation of the true positive (TP), false positive (FP), true negative (TN), and false negative (FN) used for the definition of evaluation criteria.\label{fig:criteria}}
\end{figure}

Dice score, Sensitivity, and Precision are the most commonly used evaluation criteria to assess the quality of deep medical image segmentation methods. They are defined as follows:
\begin{equation}
    \mathsf{Dice}(P,T)=\frac{2\times TP}{FP+2\times TP+FN}, 
\end{equation}
\begin{equation}
    \mathsf{Sensitivity}(P,T)=\frac{TP}{TP+FN},
\end{equation}
\begin{equation}
    \mathsf{Precision}(P,T)=\frac{TP}{TP+FP},
\end{equation}
where $TP$, $FP$, and $FN$ denote, respectively, the numbers of true positive, false positive, and false negative voxels. (See Figure \ref{fig:criteria}). The Dice score, considering the intersection region of the predicted tumor region and the actual tumor region, is a global measure of segmentation performance. It is the most popular evaluation criterion for medical image segmentation tasks. Sensitivity is the proportion, among actual tumor voxels, of voxels correctly predicted as tumors. Precision is the proportion, among predicted tumor voxels, of voxels that actually belong to the tumor region. These two criteria, thus, have to be considered jointly. A more comprehensive introduction to the evaluation criteria for medical image segmentation can be found in~\cite{taha2015metrics}.

\subsection{Limitations }
\begin{figure*}
\centering

\subfloat[]{\label{fig:demp_0}\includegraphics[width=0.5\textwidth]{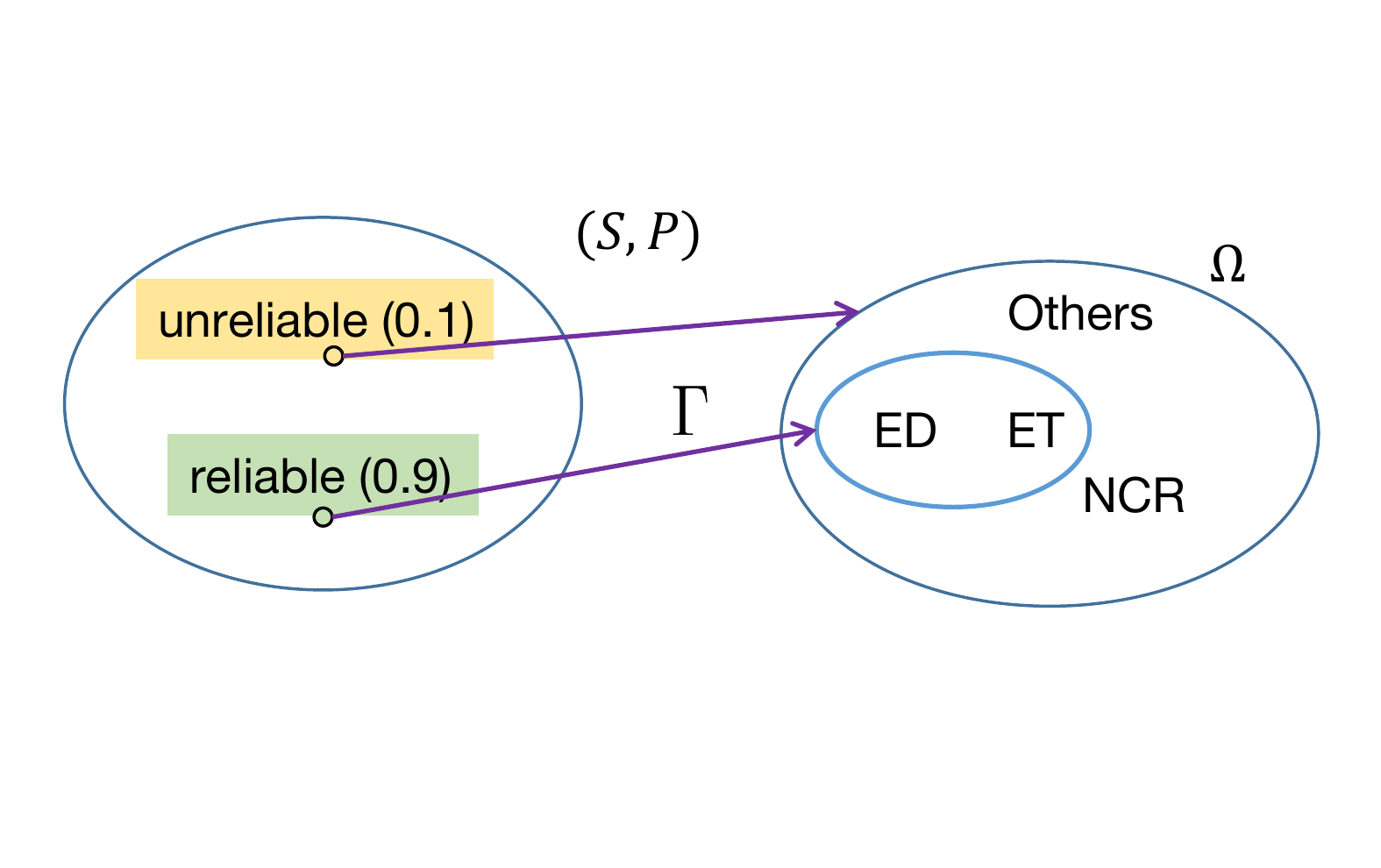}}
\subfloat[]{\label{fig:demp_1}\includegraphics[width=0.5\textwidth]{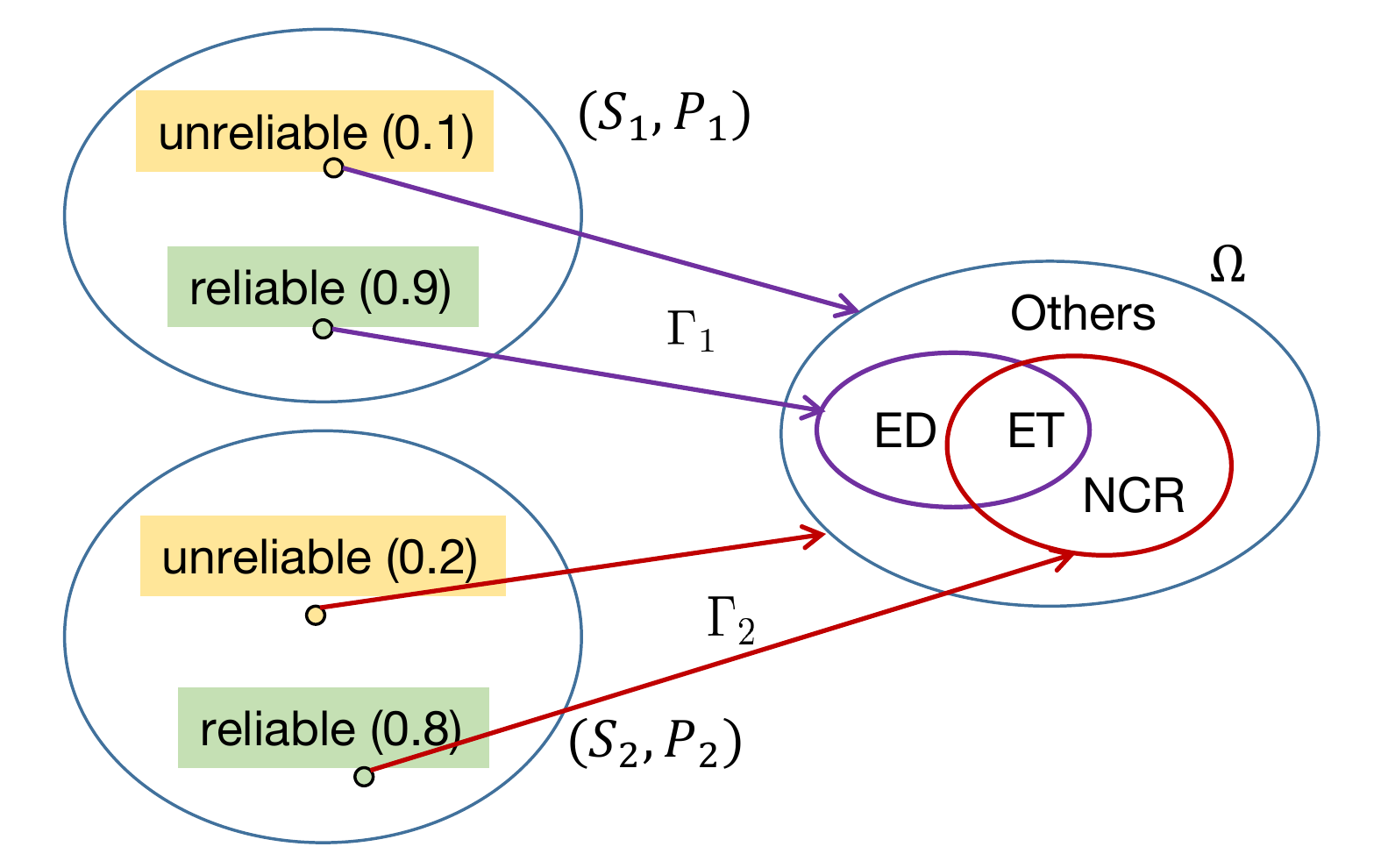}}\\
\subfloat[]{\label{fig:demp_2}\includegraphics[width=0.5\textwidth]{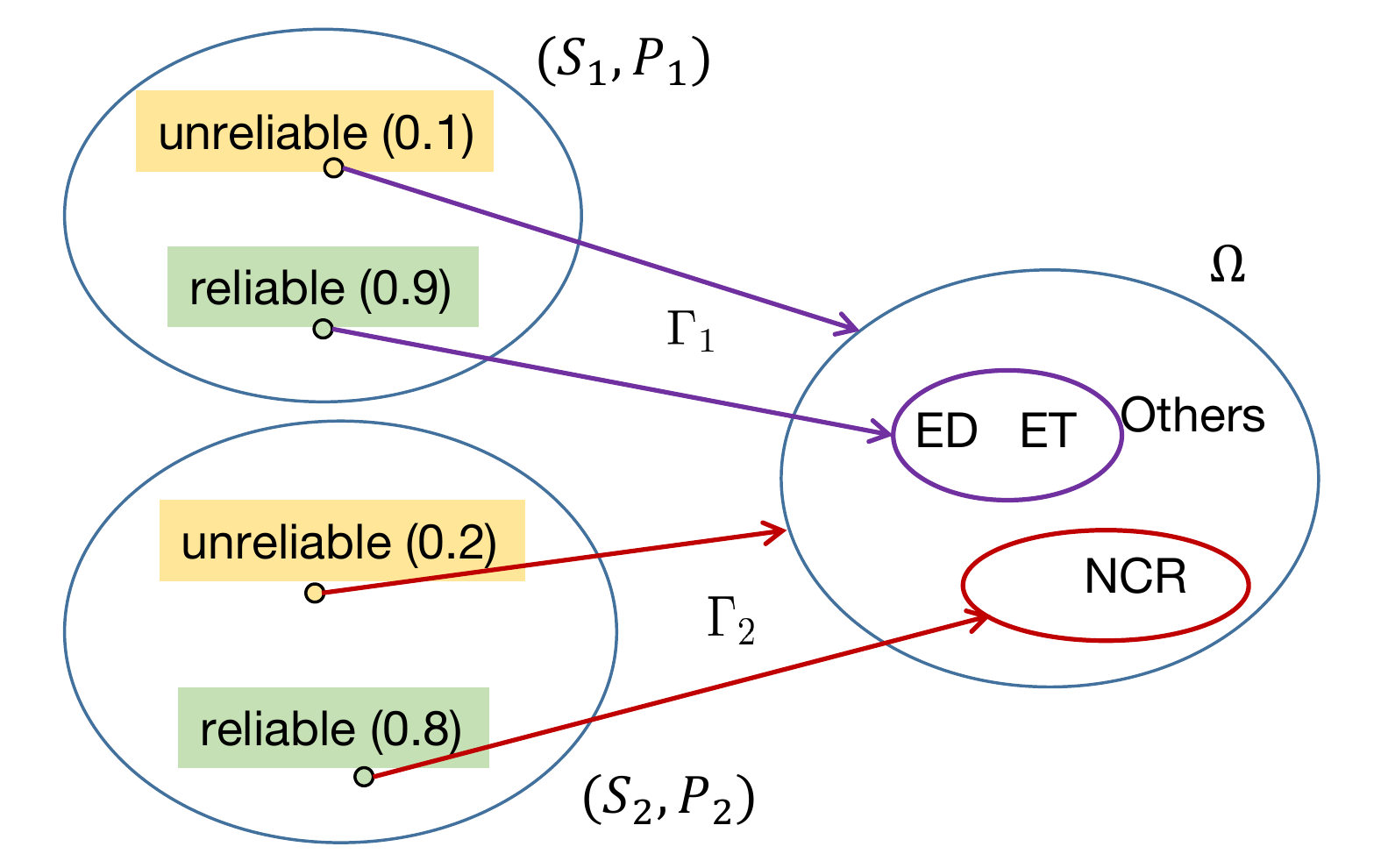}}
\caption{\label{fig:uncertainty_example}(a) Example of a segmentation task with uncertain information. (b) Example of a segmentation task with multiple sources of information (c) Example of a segmentation task with conflicting sources of information}
\end{figure*}

The above research focuses on improving the accuracy of segmentation performance under the assumption of adequate and perfect input information and accurate and appropriate prior knowledge. However, in reality, especially in the medical image segmentation domain, both the input information and prior knowledge are imperfect and contain a degree of uncertainty. Figure \ref{fig:demp_0} illustrates uncertain information taking a brain tumor segmentation task as an example. Let $X$ be the type of tumor of a voxel, and $\Omega=\{ED, ET, NCR, Others\}$, corresponding to the possibilities: edema, enhancing tumor, necrotic core, and others. Let us assume that a specialist provides the information $X \in \{ED, ET\}$, but there is a probability $p = 0.1$ that the information is unreliable. How to represent this situation by a probability function is a challenging problem. Another situation is when we have multiple information sources tainted with uncertainty, as illustrated in Figure \ref{fig:demp_1}; how can we model that kind of uncertainty and fuse the evidence? Furthermore, if the  information sources are in conflict and contain uncertainty as well, i.e., Figure~\ref{fig:demp_2}, it is difficult to represent and summarize that information by probabilistic models. Thanks to BFT, these challenges can be addressed by designing new frameworks for modeling, reasoning, and fusing imperfect (uncertain, imprecise) information. In the next section, we will give a brief introduction to BFT.

\section{Fundamentals of belief function theory}

\label{sec:BFT}

\begin{figure*}
\includegraphics[width=\textwidth]{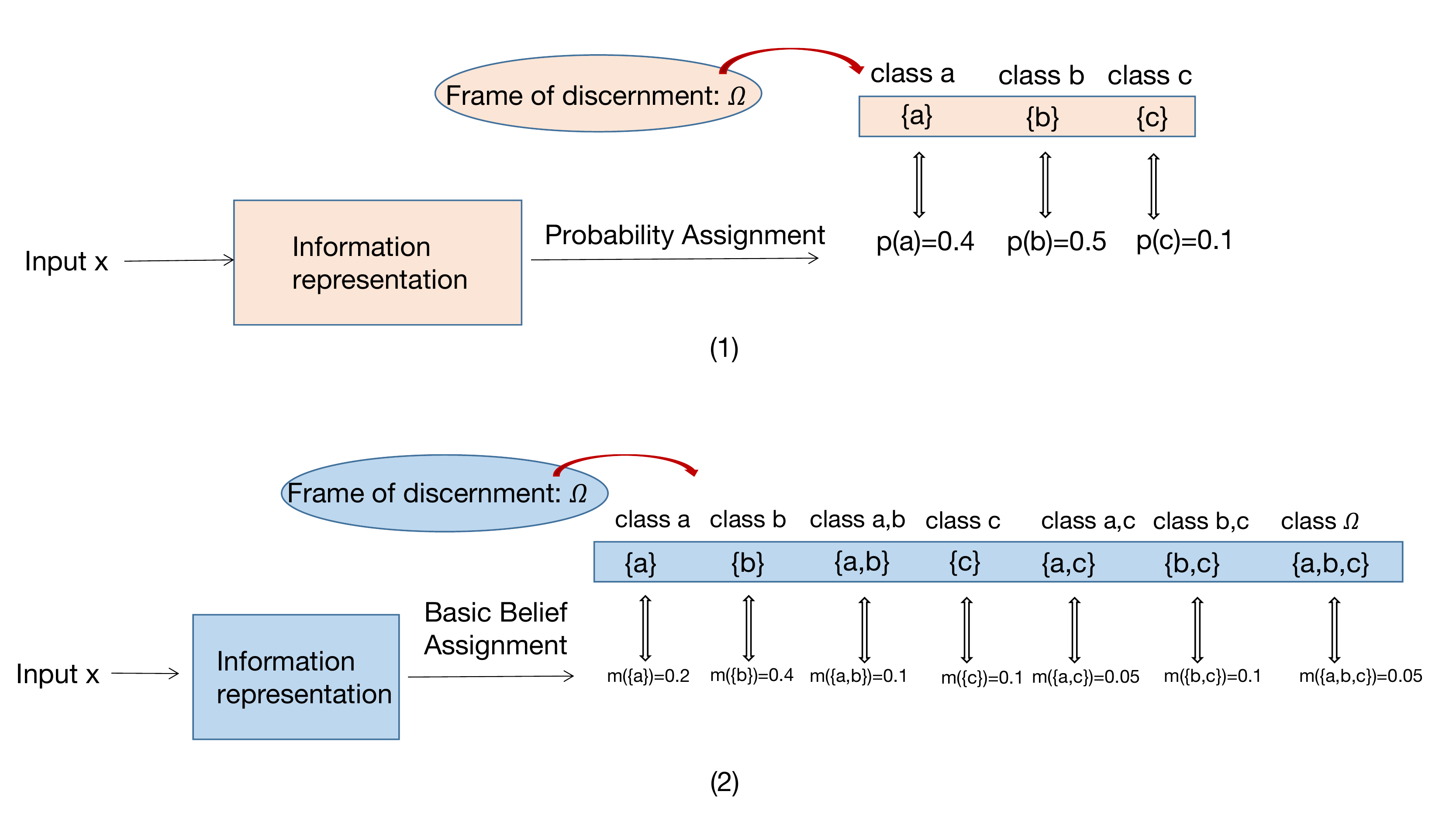}
\caption{An example of three class assignments: (1) Probabilistic model and (2) BFT model. In contrast with the probabilistic model, the BFT model can quantify uncertainty and assign it to the focal set $\{a,b\}, \{a,c\}, \{b,c\}$ and $\{a,b,c\}$ to represent its uncertainty or ignorance (here, $m$  is the evidence (mass function) about a variable $\omega$ taking values in $\Omega$, which will be introduced in Section \ref{subsec:mass}).\label{fig:probability_ds_three_class}}
\end{figure*}
BFT is a generalization of Bayesian theory, but it is more flexible than the Bayesian approach and suitable under weaker conditions~\cite{sun2018multi}, i.e.,~imperfect (uncertain, imprecise, partial) information. Figure~\ref{fig:probability_ds_three_class} shows the difference between a probabilistic model and a BFT model when applied to a three-class classification task ($\Omega=\{a, b, c\}$). For input $x$, the probabilistic model outputs the probability that $x$ belongs to classes $a$, $b$, and $c$ as 0.4, 0.5, and 0.1, respectively. In contrast, the BFT model can represent degrees of belief that $x$ belongs specifically to any subset of $\Omega$, e.g., $\{a, b\}$, $\{b, c\}$. Compared with the probabilistic model, the BFT model has more degrees of freedom to represent  uncertainty. 

In the past decades, BFT has generated considerable interest and has had great success in diverse fields, including uncertain
reasoning~\cite{smets1990combination,yager1987dempster,dubois1988representation,denoeux2008conjunctive,xiao2020generalization},
classification~\cite{denoeux1995k,denoeux2000neural,xiao2022negation} and
clustering~\cite{denoeux2004evclus,masson2009recm}, etc. It was first originated by Dempster~\cite{dempster1967upper} in the context of statistical inference in 1968 and was later formalized by Shafer~\cite{shafer1976mathematical} as a theory of evidence in 1976. In 1986, Dubois and Prade proposed an approach to the computerized processing of uncertainty~\cite{dubois2012possibility}. In 1978, Yager proposed a new combination rule of the belief function framework~\cite{yager1987dempster}. In 1990, BFT was further popularized and developed by Smets~\cite{smets1990combination} as the 'Transferable Belief Model' with the pignistic transformation for decision making. Since then, booming developments have been made. More detailed information about the development of BFT in 40~years can be found in~\cite{denzux201640}.

We will first briefly introduce the basic notions of BFT in  Section \ref{subsec:mass}, which includes evidence representation (mass functions, belief, and plausibility functions). Second, we introduce Dempster's combination rule to explain the operations of multiple sources of evidence in Section \ref{subsec:fusion}. Third, we introduce the discounting operation for unreliable sources in Section \ref{subsec:discounting}. Fourth, we introduce some commonly used decision-making methods in Section \ref{subsec:dm}. 

\subsection{Representation of evidence}
\label{subsec:mass}

Let $\Omega =\{\omega _{1},\omega _{2}, \ldots, \omega_{C}\} $ be a finite set of all possible hypotheses about some problem, called the frame of discernment. Evidence about a variable $\omega$ taking values in $\Omega$ can be represented by a mass function $m$, from the power set $2^{\Omega}$ to $[0, 1]$, such that
\begin{subequations}
\begin{align}
     \sum _{A\subseteq \Omega }m(A)&=1,\\
     m(\emptyset)&=0.
    \label{eq:evidence}
\end{align}    
\end{subequations}
 Mapping $m$ can also be called basic belief assignment (BBA). The methods used in medical image segmentation to generate mass functions will be introduced in  Section \ref{sec:mass_finction}. Each subset $A \subseteq \Omega$ such that $m(A)>0$ is called a focal set of $m$. The mass $m(\Omega)$ represents the degree of ignorance about the problem. If all focal sets are singletons, then $m$ is said to be Bayesian. It is equivalent to a probability distribution. The information provided by a mass function $m$ can be represented by a belief function $Bel$ or a plausibility function $Pl$ from $2^{\Omega }$ to $[0,1]$ defined, respectively, as
\begin{equation}
   Bel(A) = \sum _{ B\subseteq A}m(B)
   \label{eq:bel}
\end{equation}
and
\begin{equation}
   Pl(A) = \sum _{B\cap A\neq \emptyset }m(B)=1-Bel(\bar{A}),
   \label{eq:pl}
\end{equation}
for all $A\subseteq \Omega$. The quantity $Bel(A)$ can be interpreted as a degree of support to $A$, while $Pl(A)$ can be interpreted as a measure of lack of support given to the complement of $A$. 

\subsection{Dempster's rule}
\label{subsec:fusion}
In BFT, the belief about a certain question is elaborated by aggregating different belief functions over the same frame of discernment. Given two mass functions $m_{1}$ and $m_{2}$ derived from two independent items of evidence, the
final belief that supports $A$ can be obtained by combining $m_{1}$
and $m_{2}$ with Dempster's rule~\cite{shafer1976mathematical}
defined as
\begin{equation}
    (m_{1}\oplus m_{2})(A)=\frac{1}{1-\kappa }\sum _{B\cap D=A}m_{1}(B)m_{2}(D),
    \label{eq:demp1}
\end{equation}
for all $A\subseteq \Omega, A\neq \emptyset$, and $(m_{1}\oplus m_{2})(\emptyset)=0$. The coefficient $\kappa$ is the degree of conflict between $m_{1}$ and $m_{2}$, it is defined as
\begin{equation}
    \kappa=\sum _{B\cap D=\emptyset}m_{1}(B)m_{2}(D).
    \label{eq:demp2}
\end{equation}

\subsection{Discounting}
\label{subsec:discounting}
In \eqref{eq:demp2}, if $m_{1}$ and $m_{2}$ are logically contradictory, we cannot use Dempster's rule to combine them. Discounting strategies can be used to combine highly conflicting evidence~\cite{shafer1976mathematical,mercier2008refined,denoeux2019new}.
Let $m$ be a mass function on $\Omega$ and $\xi$ a coefficient
in $[0,1]$. The \emph{discounting} operation~\cite{shafer1976mathematical}
with the discount rate $\xi$ transforms $m$ into a weaker, less informative mass function defined as follows:
\begin{subequations}
\begin{align}
^{\xi}m(A)&=(1-\xi ) m(A), \quad \forall A\subset \Omega, \\
^{\xi}m(\Omega)&=(1-\xi ) m(\Omega)+ \xi.
\end{align}
\label{eq:dis}
\end{subequations}
Coefficient $1-\xi$ represents the degree of belief that the source generating $m$ is reliable. In the medical domain, the discounting operation is widely used in multimodal evidence fusion as it makes it possible to consider source  reliability in the fusion process.

\subsection{Decision-making}
\label{subsec:dm}
After combining all the available evidence in the form of a mass function, it is necessary to make a decision. In this section, we introduce some classical BFT-based decision-making methods.

\paragraph*{Upper and lower expected utilities}

Let $u$ be a utility function. The lower and upper expectations of $u$ with respect to $m$ are defined, respectively, as the averages of the minima and the maxima of $u$ within each focal set of $m$:
\begin{subequations}
\begin{align}
\underline{E}_m(u)=\sum_{A\subseteq \Omega} m(A) \min_{\omega\in A} u(\omega),\\
\overline{E}_m(u)=\sum_{A\subseteq \Omega} m(A) \max_{\omega\in A} u(\omega).
\end{align}
\label{eq:utlity}
\end{subequations}
When $m$ is Bayesian, $\underline{E}_m(u)= \overline{E}_m(u) $. If $m$ is logical with focal set $A$, then $\underline{E}_m(u)$ and $\overline{E}_m(u)$ are, respectively, the minimum and maximum of $u$ in $A$. The lower or upper expectations can be chosen for the final decision according to the given task and decision-maker's attitude.

\paragraph*{Pignistic criterion}
In 1990, Smets proposed a pignistic transformation~\cite{smets1990combination}
that distributes each mass of belief distributed equally among the elements of $\Omega$. The pignistic probability distribution is defined as
\begin{equation}
 BetP(\omega)=\sum _{\omega\in A}\frac{m(A)}{\mid A\mid }, \quad  \forall \omega\in \Omega,  
 \label{eq:pignistic}
\end{equation}
where $\left | A \right | $ denotes the cardinality of $A \subseteq \Omega$.

Besides the above methods, there are various decision-making methods proposed
for BFT, such as Generalized OWA criterion~\cite{yager1992decision},
Generalized minimax regret~\cite{YAGER2004109}, Generalized divergence~\cite{xiao2022generalized}, etc. More details about decision-making with BFT can be found in the review paper~\cite{denoeux2019decision}.

\section{Methods to generate mass functions}
\label{sec:mass_finction}

\begin{figure*}
\includegraphics[width=\textwidth]{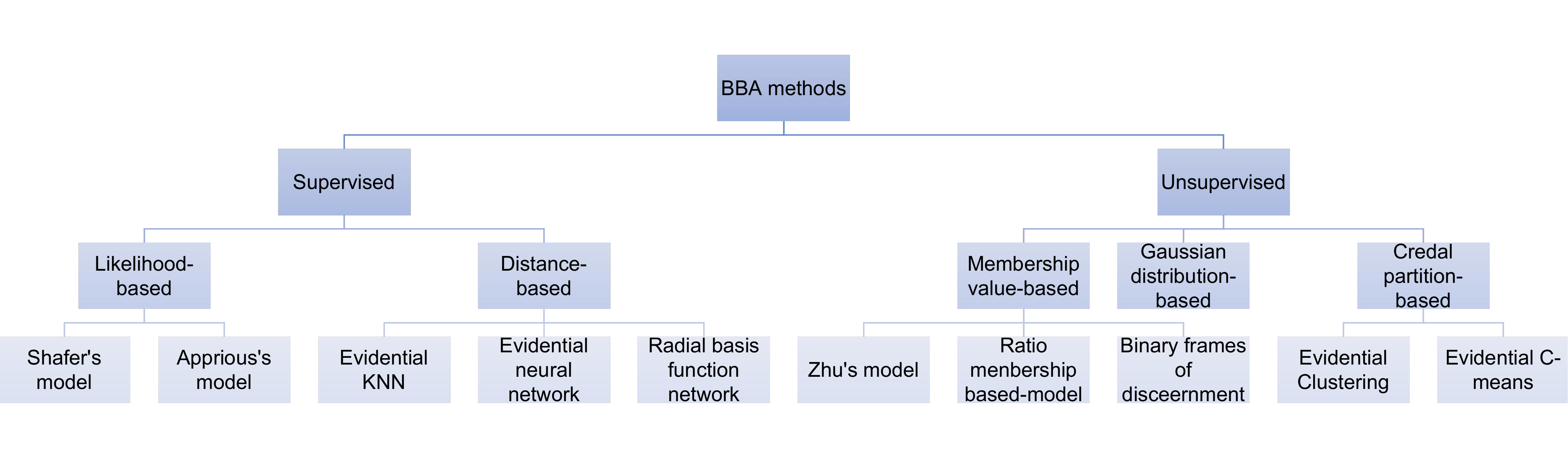}
\caption{Overview of BBA methods.\label{fig:bba}}
\end{figure*}

To model segmentation uncertainty, the first step is to generate mass functions. In this section, we introduce the BBA methods applied to medical image segmentation. Figure~\ref{fig:bba} is an overview of BBA methods. In general, those methods can be separated into supervised and unsupervised methods according to whether annotations are used to optimize the parameters of BBA models or not. 

\subsection{Supervised BBA methods}
\label{subsec:supervised}

Supervised BBA methods can be classified into two categories. One is
the likelihood-based methods, such as Shafer's
model~\cite{shafer1976mathematical} and Appriou's
model~\cite{appriou1999multisensor,appriou200501}. The other category is composed of distance-based methods, such as the evidential KNN rule~\cite{denoeux1995k}, the evidential neural network classifier~\cite{denoeux2000neural} and Radial basis function
networks~\cite{denoeux2019decision}. It should be noted that the distance-based
BBA methods can easily be merged with popular deep segmentation models and
have shown promising results~\cite{huang2021belief,huang2022lymphoma}.

\subsubsection{Likelihood-based BBA methods}
\label{subsubsection:lik_methods}

\paragraph*{Shafer's model}
In~\cite{shafer1976mathematical}, Shafer proposed a likelihood-based evidential
model to calculate mass functions. Assuming that conditional density functions
$f(x\mid \omega_c)$ are known, then the conditional likelihood associated with the
pattern $X$ is defined by $\ell (\omega_c\mid x)=f(x\mid \omega_c)$. The mass functions are defined according to the knowledge of all hypotheses $\omega_1, \ldots, \omega_C $. Firstly, the plausibility of a simple hypothesis $\omega_c$ is proportional to its
likelihood. The plausibility is thus given by
\begin{equation}
    Pl(\{\omega_c\})=\hbar \cdot \ell (\omega_c\mid x), \quad  \forall \omega_c \in \Omega,
    \label{eq:14}
\end{equation}
where $\hbar $ is a normalization factor with $\hbar=1/  \max_{\omega\in\Omega} \ell(\omega \vert x)$. The plausibility of a set $A$ is thus given by
\begin{equation}
  Pl(A)=\hbar\cdot \underset{\omega_c \in A}{\max} \ell (\omega_c\mid x).
\end{equation}

\paragraph*{Appriou's model}
 
Same as Shafer's one, Appriou~\cite{appriou1999multisensor,appriou200501}
proposed two likelihood-based models to calculate mass functions with the frame of discernment $\Omega= \{\omega_{c}, \neg{\omega_{c}} \}$. For the first model, the mass functions are defined by
\begin{subequations}
\begin{align}
m(\{\omega_{c}\})&=0,\\
m(\{\neg{\omega_{c}}\})&=\alpha_{c}  ({1-\hbar \cdot \ell (\omega_{c}\mid x) }) ,\\
m(\Omega)&=1-\alpha_{c}{ (1-\hbar \cdot \ell (\omega_{c}\mid x ))},
    \label{eq:15}
\end{align}
\end{subequations}
where $\alpha_{c}$ is a reliability factor depending on the hypothesis $\omega_{c}$ and on the source information. The second model is defined as
\begin{subequations}
\label{eq:appriou2}
\begin{align}
m(\{\omega_{c}\})&=\alpha_{c} \cdot \hbar \cdot \ell (\omega_{c}\mid x)/  {(1+\hbar \cdot \ell (\omega_{c}\mid x))}, \\
m(\{\neg{\omega_{c}}\})&=\alpha_{c}/ {(1+\hbar \cdot \ell (\omega_{c}\mid x))} ,\\
m(\Omega)&=1-\alpha_{c}.
    \label{eq:16}
\end{align}
\end{subequations}

\subsubsection{Distance-based BBA methods}

\paragraph*{Evidential KNN ({EKNN}) rule}
 
In~\cite{denoeux1995k}, Den{\oe}ux proposed a distance-based KNN classifier for classification tasks. Let $N_K(x)$ denote the set of the $K$ nearest neighbors of $x$ in learning set $Z$. Each $x_{i}\in N_K(x)$ is considered as a piece of evidence regarding the class label of $x$. The strength of evidence decreases with the distance between
$x$ and $x_{i}$. The evidence of $(x_{i},y_{i})$ support class $c$ is represented by
\begin{subequations}
\begin{align}
m_i(\{\omega_{c}\})&=\varphi _{c}(d_{i})y_{ic}, \quad  1 \le c \le C,\\
m_{i}(\Omega)&=1-\varphi_{c}(d_{i}),
\end{align}
\label{eq:18}
\end{subequations}
where $d_{i}$ is the distance between $x$ and $x_{i}$, which can be Euclidean or other distance function; and $y_{ic}=1$ if $y_{i}=\omega_{c}$ and $y_{ic}=0$ otherwise. Function $\varphi _{c}$ is defined as
\begin{equation}
    \varphi _{c}(d)=\alpha \exp(-\gamma d^{2}),
    \label{eq:19}
\end{equation}
where $\alpha$ and $\gamma$ are two tuning parameters. The evidence of the $K$ nearest neighbors of $x$ is fused by Dempster's rule:
\begin{equation}
m=\bigoplus _{x_{i}\in N_K(x)}m_{i}.  
\label{eq:20}
\end{equation}

The final decision is made according to maximum plausibility. The detailed optimization of these parameters is described in~\cite{zouhal1998evidence}. Based on this first work, Den{\oe}ux et al. proposed the contextual discounting evidential KNN rule~\cite{denoeux2019new} with partially supervised learning to address the annotation limitation problem.

\paragraph*{Evidential neural network (ENN)}

The success of machine learning encouraged the exploration of applying belief function theory with learning methods. In~\cite{denoeux2000neural}, Den{\oe}ux proposed an ENN classifier in which mass functions are computed based on distances to prototypes. 

The ENN classifier is composed of an input layer of $H$ neurons, two hidden layers, and an output layer. The first input layer is composed of $I$ units, whose weights vectors are prototypes $p_1,\ldots, p_I$ in input space. The activation of unit $i$ in the prototype layer is
\begin{equation}
    s_i=\alpha _i \exp(-\gamma_i d_i^2),   
    \label{eq:si}
\end{equation}
where $d_i= \left |  x-p_i \right |  $ is the Euclidean distance between input vector $x$ and prototype $p_i$, $\gamma_i>0$ is a scale parameter,  and $\alpha_i \in [0,1]$ is an additional parameter. The second hidden layer computes mass functions $m_i$ representing the evidence of each prototype $p_i$, using the following equations: 
\begin{subequations}
\begin{align}
m_i(\{\omega _{c}\})&=u_{ic}s_i, \quad c=1, \ldots , C\\
m_{i}(\Omega)&=1-s_i, 
\label{eq:9}
\end{align}
\end{subequations}
where $u_{ic}$ is the membership degree of prototype $i$ to class $\omega_c$, and $\sum _{c=1}^C u_{ic}=1$. Finally, the third layer combines the $I$ mass functions $m_1,\ldots,m_I$ using Dempster's rule. The output mass function $m=\bigoplus_{i=1}^I m_i$ is a discounted Bayesian mass function that summarizes the evidence of the $I$ prototypes.

\paragraph*{Radial basis function (RBF) network}
In~\cite{huang2022lymphoma}, Huang et al. confirmed that RBF network can be an alternative approach to ENN  based on the aggregation of weights of evidence. Consider an RBF network with $I$ prototype (hidden) units. The activation of hidden unit $i$ is
\begin{equation}
    s_i=\exp(-\gamma_i d_i^2),   
    \label{eq:activRBF}
\end{equation}
where, $d_i=\left |  \bx-\bp_i \right |   $ is the Euclidean distance between input vector $\bx$ and prototype $\bp_i$, and $\gamma_i>0$ is a scale parameter. Here we only show an example of binary classification task $C=2$ and $\Omega=\{\omega_1,\omega_2\}$
(The multi-class example can be found in~\cite{denoeux19d}). Let $v_{i}$ be the weight of the connection between hidden unit $i$ and the output unit, and let $w_i=s_i v_i$ be the product of the output of unit $i$ and weight $v_i$. The quantities $w_i$ can be interpreted as weights of evidence for class $\omega_1$ or $\omega_2$, depending on the sign of $v_i$. To each prototype $i$ can be associated with the following simple mass function:
\[m_{i}=\{\omega_1\}^{w_{i}^+} \oplus \somega{2}^{w_{i}^-},\]
where $\{\omega\}^w$ is the notation for the simple mass function focussed on $\{\omega\}$ with weight of evidence $w$, and $w_{i}^+$, $w_{i}^-$ denote, respectively, the positive and negative parts of $w_i$. Combining the evidence of all prototypes in favor of $\omega_1$ or $\omega_2$ by Dempster's rule, we get the mass function $m=\bigoplus_{i=1}^I m_{i}$ with the following expression:
\begin{subequations}
\label{eq:m12}
\begin{align}
m(\somega{1}) &= \frac{[1-\exp(-w^+)]\exp(-w^-)}{1-\kappa}, \\
m(\somega{2}) &= \frac{[1-\exp(-w^-)]\exp(-w^+)}{1-\kappa},\\
m(\Omega) &= \frac{\exp(-w^+-w^-)}{1-\kappa}=\frac{\exp(-\sum_{i=1}^I \left |w_i \right | )}{1-\kappa},
\end{align}
\end{subequations}
where $\kappa$ is the degree of conflict between $\{\omega_1\}^{w^+}$ and $\{\omega_2\}^{w^-}$ given by
\begin{equation}
\label{eq:conf12}
\kappa=[1-\exp(-w^+)] [1-\exp(-w^-)].
\end{equation}

\subsection{Unsupervised BBA methods}
\label{subsec:unsupervised}

The goal of unsupervised BBA methods is to generate mass functions without any label information. In earlier BBA studies, Fuzzy C-means (FCM)~\cite{dunn1973fuzzy} was the most popular algorithm used to generate membership values (MVs). Based on MVs, the authors can obtain mass functions according to some domain knowledge, e.g.,~threshold~\cite{zhu2002automatic}, or user-specific parameters~\cite{ghasemi2012brain}. The sigmoid and  one-sided Gaussian function
can also be used to generate MVs~\cite{safranek1990evidence}. The notion of credal
partition~\cite{denoeux2004evclus}, an extension of fuzzy partition, enables us to generate mass functions directly~\cite{masson2008ecm}. Besides these two popular BBA methods, mass functions can also be generated from the Gaussian distribution of the input to the cluster center~\cite{chen2012manifold}.

\subsubsection{MVs-based BBA methods}
\label{subsub:MVs}
\paragraph*{FCM}

Considering that there are some BFT-based methods that use FCM to generate MVs,  we introduce it here to offer a basic view for readers. With FCM, any $\mathbf{x}$ has a set of coefficients $w_{k}(x)$ representing the degree of membership in the $k$th cluster. The centroid of a cluster is the mean of all points, weighted by the $m$-th power of their membership degree,
\begin{equation}
c_{k}=\frac{{\sum _{x}{w_{k}(x)}^{m}x}}{ {\sum _{x}{w_{k}(x)}^{m}}},
\end{equation}
where $m$ is the hyper-parameter that controls how fuzzy the cluster will be. The higher it is, the fuzzier. Given a finite set of data, the FCM algorithm returns a list of cluster centers $P=\{c_{1}, \ldots, c_{C}\}$ and a partition matrix $W=(\omega_{ij}), i=1,\ldots, N, j=1, \ldots, C$,

\begin{equation}
    w_{{ij}}={\frac  {1}{\sum _{{k=1}}^{{C}}\left({\frac  {\left\|{\mathbf  {x}}_{i}-{\mathbf  {c}}_{j}\right\|}{\left\|{\mathbf  {x}}_{i}-{\mathbf  {c}}_{c}\right\|}}\right)^{{{\frac  {2}{m-1}}}}}},
\end{equation}
where $w_{ij}$, is the degree of membership of $\mathbf{x}_{i}$  to cluster $\mathbf{c}_{j}$. The objective function is defined as
\begin{equation}
    \underset{P }{arg \max} \sum_{i=1}^{N} \sum_{j=1}^{C} w_{ij}^m \left\|  \mathbf{x}_i-\mathbf{c}_j  \right\| . 
\end{equation}

\paragraph*{Zhu's model}


\begin{figure*}

\subfloat[]{\label{fig:overlaping}\includegraphics[width=0.5\textwidth]{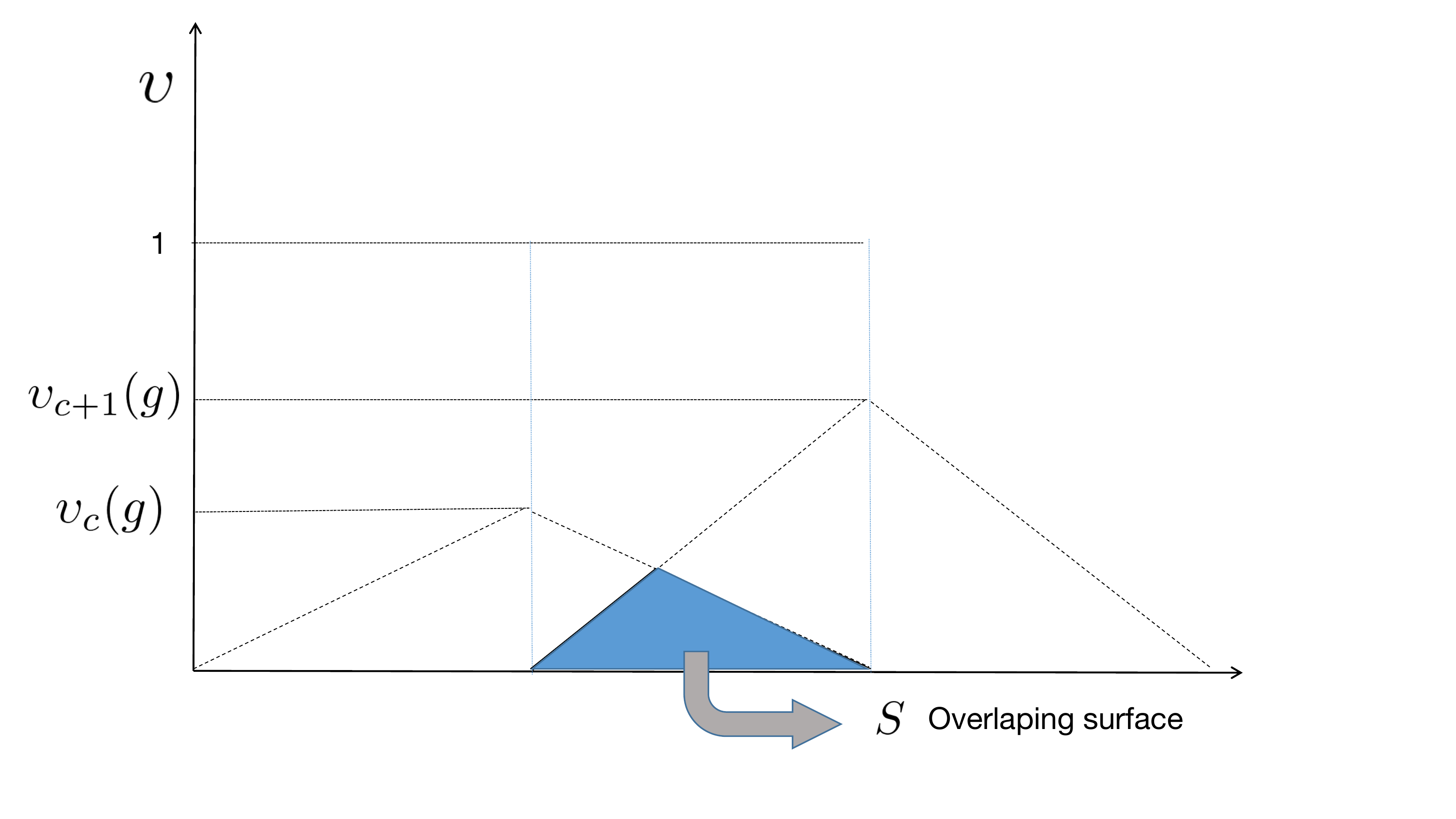}}
\subfloat[]{\label{fig:maximumambigulity}\includegraphics[width=0.5\textwidth]{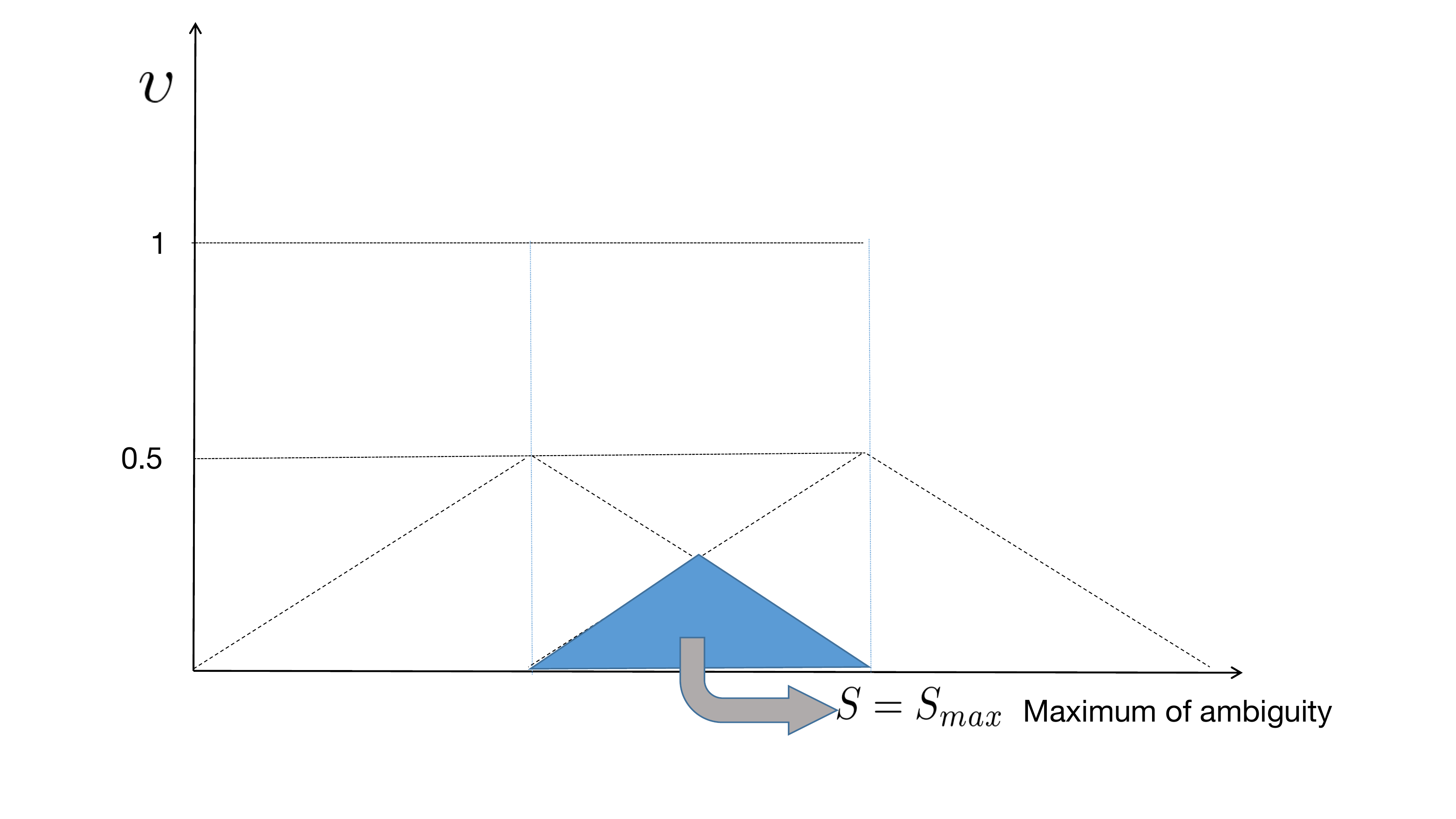}}
\caption{\label{fig:zhu}(a) Construction of triangular membership functions and (b) maximum ambiguity case: $\upsilon _c(g)=\upsilon _{c+1}(g)=0.5$. 
}
\end{figure*}

In~\cite{zhu2002automatic}, Zhu et al. proposed a method to determine mass
functions using FCM and neighborhood information. The mass assigned to simple hypothesis
$\{ \omega_c \}$ is directly obtained from the filtered membership functions
$\upsilon _{c}(g)$ of the gray level $g(x,y)$ to cluster $c$ as
$m(\{\omega_{c}\}) = \upsilon_{c}(g)$. For a given gray level, the piece of evidence of
belonging to the cluster $c$ is, thus, directly given by its degree of
membership to the same cluster. If there is a high ambiguity in assigning a
gray level $g(x,y)$ to cluster $c$ or $c+1$, that is, $\left | \upsilon _c(g)-\upsilon _{c+1}(g) \right | < \varepsilon $, where $\varepsilon $ is a thresholding value, then a double hypothesis is formed. The value of the threshold $\varepsilon $ is chosen
depending on the application. The authors suggested fixing
$\varepsilon $ at 0.1. Once the double hypotheses are formed, their associated
mass is calculated according to the following formula:
\begin{equation}
   m(\{\omega_{c},\omega_{c+1}\})
   =\frac{S[\upsilon _{c}(g), \upsilon _{c+1}(g)]} { 2S_{max}}, 
   \label{eq:29}
\end{equation}
where $S$ represents the surface of a triangle and $S_{max}$ is the maximum of ambiguity. The surface of such a triangle depends both on the degrees of the membership functions of $g(x,y)$ to clusters $c$ and $c+1$ and on the conflicts between these MVs. Figure~\ref{fig:overlaping} shows how the triangle is constructed and how the mass of double hypotheses $\{\omega_c, \omega_{c+1} \}$ is derived from the surface of the triangle. The vertical axis of  Figure~\ref{fig:overlaping} represents the MVs. The surfaces of the two dotted triangles define two so-called triangular membership functions corresponding to classes $c$ and $c+1$. The two triangles are isosceles and have the same length for their bases. The heights of the triangles are equal to $m(\{\omega_c\})$ and $m(\{\omega_{c+1}\})$, respectively. The overlapping surface $S$ of the two triangles represents the MV to the double hypothesis $\{\omega_c, \omega_{c+1} \}$. Therefore, the mass value attributed to the double hypothesis $\{\omega_c, \omega_{c+1} \}$ can be directly calculated from the surface $S$.  Figure~\ref{fig:maximumambigulity} shows the condition of the maximum ambiguity case.

\paragraph*{Ratio MV (RMV) transformation}

In~\cite{ghasemi2012brain}, Ghasemi et al. proposed a ratio membership value transformation method to calculate mass functions. The FCM algorithm was first used to generate MVs $f_{\omega_c}$ for each pixel. Then the MVs are used to build the mass functions. For this purpose, the three ratios of the available MVs are calculated, corresponding to three situations: no-uncertainty (NU), semi-uncertainty (SU), and perfect-uncertainty (PU). First, PU is a critical situation in which the RMVs are smaller than $\alpha$, then the mass function is calculated as $m(\{\omega_1\})=m (\{ \omega_2 \})=m (\Omega )=(f_{\omega_1}+f_{\omega_2})/3$. Second, two thresholds $\alpha$ and
$\beta$ with $\alpha=1.5$ and $\beta=3$ are selected to control the boundary between SU and PU, and between NU and SU, separately. For example, with 
\[
f_{\omega_1}=0.18,\quad
f_{\omega_2}=0.81,\quad
RMV=f_{\omega_1}/f_{\omega_2}= 4.5, \quad RMV>\beta,\]
the two MVs fall in the NU category. If
\[f_{\omega_1}=0.25,\quad f_{\omega_2}=0.65,\quad RMV=f_{\omega_1}/f_{\omega_2}=2.6,\quad \alpha<RMV<\beta,\]
the two MVs are in the SU category. The mass functions are calculated as
\begin{subequations}
\begin{align}
 m(\{\omega_1\})&=f_{\omega_1}-\frac{\lambda_{\omega_1,\omega_2}}{2},\\
 m(\{\omega_2\})&=f_{\omega_2}-\frac{\lambda_{\omega_1,\omega_2}}{2},\\
 m (\Omega)&=\lambda_{\omega_1,\omega_2},
\end{align} 
\label{eq:31}
\end{subequations}
where $\lambda$ is an uncertainty distance value  defined as $\lambda_{\omega_1, \omega_2}=\frac{\left | f_{\omega_1}-f_{\omega_2} \right |}{\beta-\alpha}$.

\subsubsection{Evidential C-means (ECM)}
In~\cite{denoeux2004evclus}, Den{\oe}ux et al. proposed an evidential clustering
algorithm, called EVCLUS, based on the notion of credal partition,
which extends the existing concepts of hard, fuzzy (probabilistic), and
possibilistic partition by allocating to each object a ``mass of belief'', not only
to single clusters but also to any subsets of $\Omega= \{\omega_{1}, \ldots , \omega_{C}\}$. 

\paragraph*{Credal partition}
 Assuming there is a collection of five objects for two classes, mass functions for each source are given in Table \ref{tab:credal partition}. They represent different situations: the mass function of object 1 indicates strong evidence that the class of object 1 does not lie in $\Omega$; the class of object 2 is completely unknown, and the class of object 3 is known with certainty; the cases of objects 4 and 5 correspond to situations of partial knowledge ($m_{5}$ is Bayesian). The EVCLUS algorithm generates a credal partition for dissimilarity data by minimizing a cost function.

\begin{table}
    \centering
    \caption{Example of credal partition}
    \begin{tabular}{|c|c|c|c|c|c|}
    \hline
    $A$& $m_{1}(A)$&$m_{2}(A)$&$m_{3}(A)$&$m_{4}(A) $&$m_{5}(A) $\\
    \hline
    $\{ \emptyset \}$&1&0&0&0&0 \\
    $\{a\}$&0&0&1&0.5&0.6\\
    $\{b\}$&0&0&0&0.3&0.4\\
    $\{a,b\}$&0&1&0&0.2&0\\
    \hline
    \end{tabular}
    \label{tab:credal partition}
\end{table}

\paragraph*{Evidential C-Means (ECM)}

The ECM algorithm~\cite{masson2008ecm} is another method for generating a credal partition from data. In ECM, a cluster is represented by a prototype $p_{c}$. For each non-empty set $A_{j}\subseteq\Omega$, a prototype $\bar{p_{j}}$ is defined as the center of mass of the prototypes $p_c$ such that $\omega_c\in A_j$. Then the non-empty focal set is defined as $F=\{A_{1}, \ldots , A_{f}\}\subseteq 2^{\Omega}\setminus\left\{\emptyset\right \} $. Deriving a credal partition from object data implies determining,
for each object $x_i$, the quantities $m_{ij}=m_i(A_j), A_i\ne \emptyset, A_j \subseteq \Omega$, in such a way that $m_{ij}$ is low (resp. high) when the distance between $x_i$ and the focal set $\bar{p_{j}}$ is high (resp. low). The distance between an object and
any nonempty subset of $\Omega$ is then defined by
\begin{equation}
    d_{ij}^2=\left\|  x_{i}-\bar{p_{j}}\right\| ^2.
    \label{eq:27}
\end{equation}

\subsubsection{Gaussian distribution (GD)-based model}

Besides the FCM-based and credal partition-based BBA method, the mass functions
can also be generated from GD~\cite{chen2012manifold}. The mass of simple hypotheses $\{\omega_c\}$ can be obtained from the assumption of GD according to the information $x_i$ of a pixel from an input image to cluster $c$ as follows:
\begin{equation}
m(\{\omega_c\})=\frac{1}{\sigma_c \sqrt{2\pi}} \exp{ \frac{-(x_i-\mu_{c})^2}{2 \sigma_{c}^{2}}},
\label{eq:32}
\end{equation}
where $\mu_c$ and $\sigma^2_c$ represent, respectively, the mean and the variance of the cluster $c$, which can be estimated by
\begin{equation}
    \mu_c=\frac{1}{n_c}\sum_{i=1}^{n_c}x_i,
    \label{eq:33}
\end{equation}
\begin{equation}
\sigma _c^2=\frac{1}{n_c}\sum_{i=1}^{n_c}(x_i-\mu_c)^2,
\label{eq:34}
\end{equation}
where $n_c$ is the number of pixels in the cluster $c$. The mass of multiple hypotheses  $\{ \omega_1, \omega_2, \ldots , \omega_T \}$ is determined as
\begin{equation}
    m(\{ \omega_1, \omega_2, \ldots , \omega_T\})=\frac{1}{\sigma_t\sqrt{2\pi}} \exp{\frac{-(x_i-\mu_{t})^2}{2\sigma_t^{2}}},
    \label{eq:35}
\end{equation}
where $\mu_t=\frac{1}{T} \sum_{i=1}^{T} \mu_i$, $\sigma_t=\max(\sigma_1, \sigma_2, \ldots ,\sigma_T), 2 \le T \le C$, $C$ is the number of clusters.

\subsubsection{Binary frames of Discernment (BFOD)}
 
Under the assumption that the membership value is available, Safranek et al. introduced a BFOD-based BBA method~\cite{safranek1990evidence} to transform membership values into mass functions. The BFOD is constructed as $\Omega = \{ \omega, \neg \omega \}$
with a function $cf(\nu)$, taking values in $\left [ 0,1 \right ] $ that assigns confidence factors. The sigmoid and one-sided Gaussian functions are the most appropriate functions for defining $cf(\nu)$ according to the authors. Once a confidence value is obtained, the transformation into mass functions can be accomplished by defining
appropriate transfer functions:
\begin{subequations}
\begin{align}
\label{eq:momega}
  m(\{\omega\})&=\frac{B}{1-A}cf(\nu)-\frac{AB}{1-A},\\
\label{eq:mnegomega}
 m(\{\neg \omega\})&=\frac{-B}{1-A}cf(\nu)+B,\\
 m(\Omega)&=1-m(\{\omega\})-m(\{\neg \omega\}),
\end{align}
\label{eq:28}
\end{subequations}
where $A$ and $B$ are user-specific parameters. In \eqref{eq:momega} and \eqref{eq:mnegomega}, the left-hand side stays clamped at zero when the right-hand side goes negative. The parameter $A$ is the confidence-factor axis intercept of the curve that depicts the dependence of $m(\{ \omega\})$ on confidence factors, and $B$ is the maximum support value assigned to $m(\{ \omega\})$ or $m(\{ \neg \omega\})$.

\subsection{Discussion}

If segmentation labels are available, we suggest using the supervised BBA
methods to generate mass functions, especially the distance-based BBA methods,
because these methods can be merged easily with the popular deep learning methods and construct a learnable end-to-end deep segmentation model. An example of this approach can be found in~\cite{huang2022lymphoma}, where the authors first apply the idea of merging the ENN or RBF model with a deep medical image segmentation model by mapping feature vectors into mass functions instead of using the softmax transformation to map them into probabilities.

If the segmentation labels are not available or only partially available, we can still generate mass functions using  unsupervised BBA methods, e.g.,~using ECM for tumor segmentation~\cite{lian2018unsupervised}. To our best knowledge, there is no research  on combining unsupervised BBA methods with deep segmentation models even though some studies already make the end-to-end neural network-based evidential clustering model possible, e.g.,~\cite{denoeux2021nn}. We believe it is a good choice for researchers who work on unlabeled datasets and wish to directly quantify segmentation uncertainty, they can consider applying unsupervised BBA methods to deep segmentation models.

\section{BFT-based medical image segmentation methods}
\label{sec:BFT-based fusion}

\begin{figure*}
\subfloat[]{\label{fig:composition_modality}\includegraphics[width=0.5\textwidth]{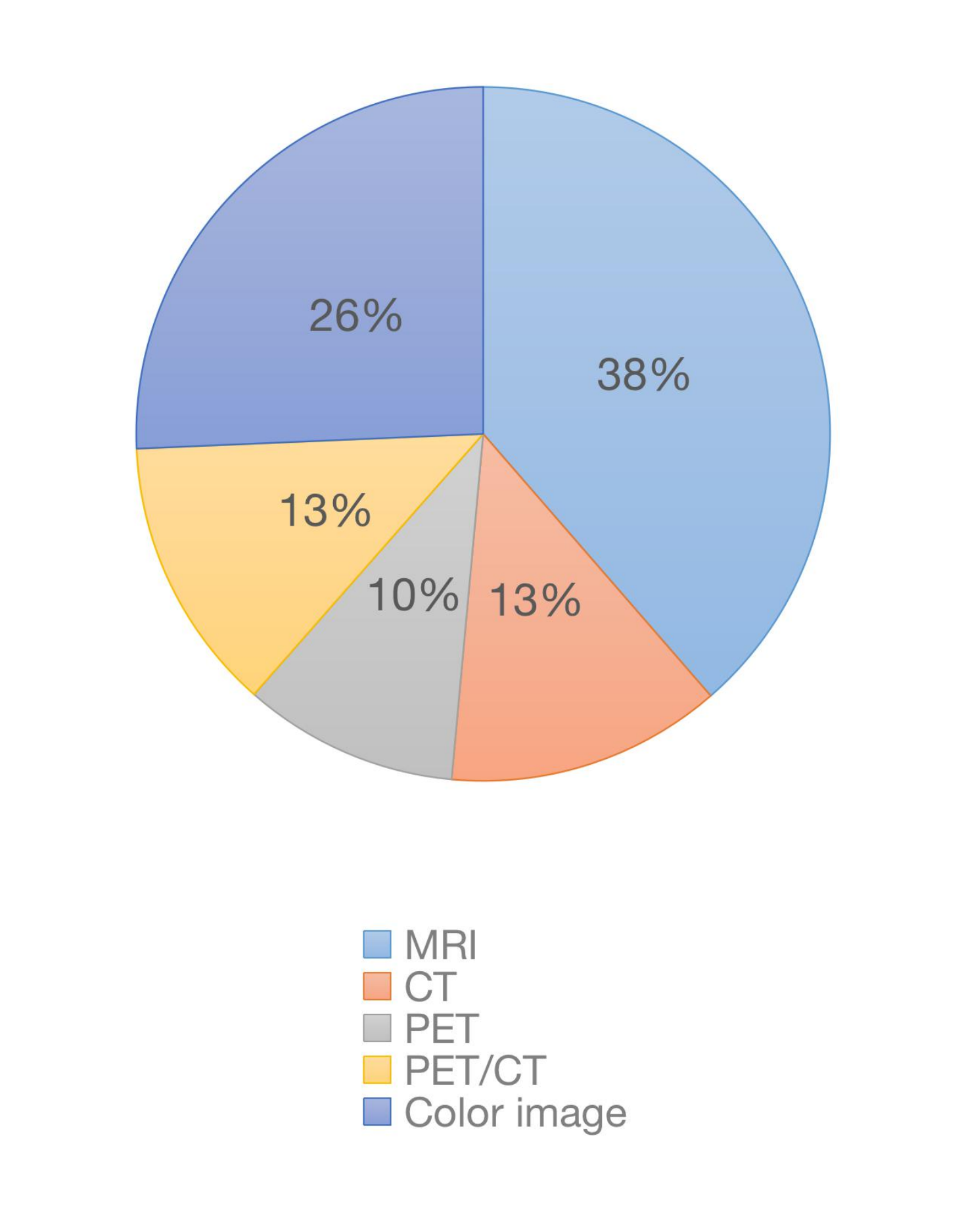}}
\subfloat[]{\label{fig:compsition_app}\includegraphics[width=0.5\textwidth]{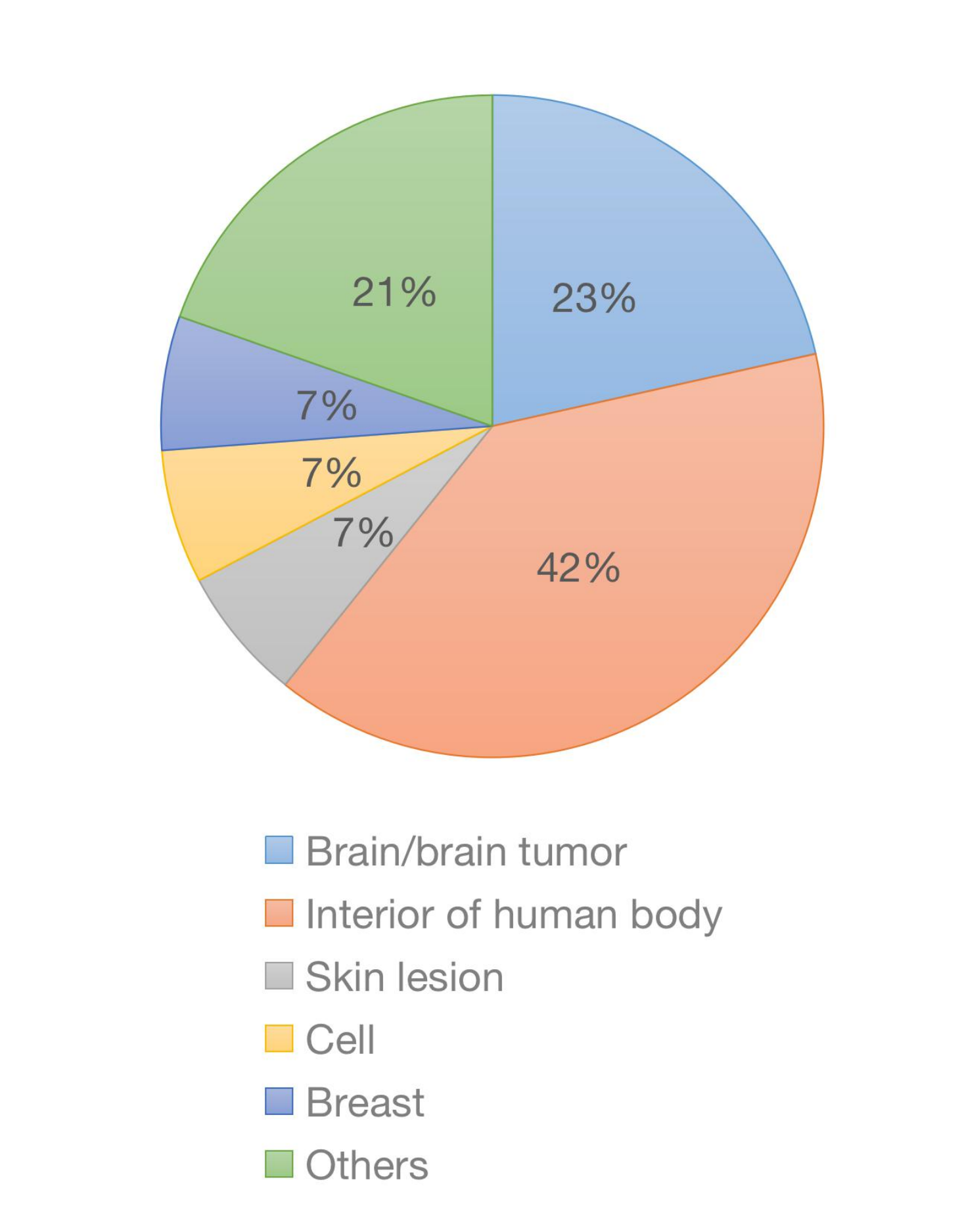}}
\caption{\label{fig:compositation}(a) Distribution of input modalities of  medical images, (b) Distribution of specific medical applications.}
\end{figure*}

To summarize the BFT-based medical image segmentation methods, we can either classify them by the input modality of the images or by the specific clinical application. Figure~\ref{fig:composition_modality} shows the proportion of types of medical images applied in the segmentation task and Figure~\ref{fig:compsition_app} displays the proportions of application in the medical image segmentation task. 

\begin{figure*}
\includegraphics[width=\textwidth]{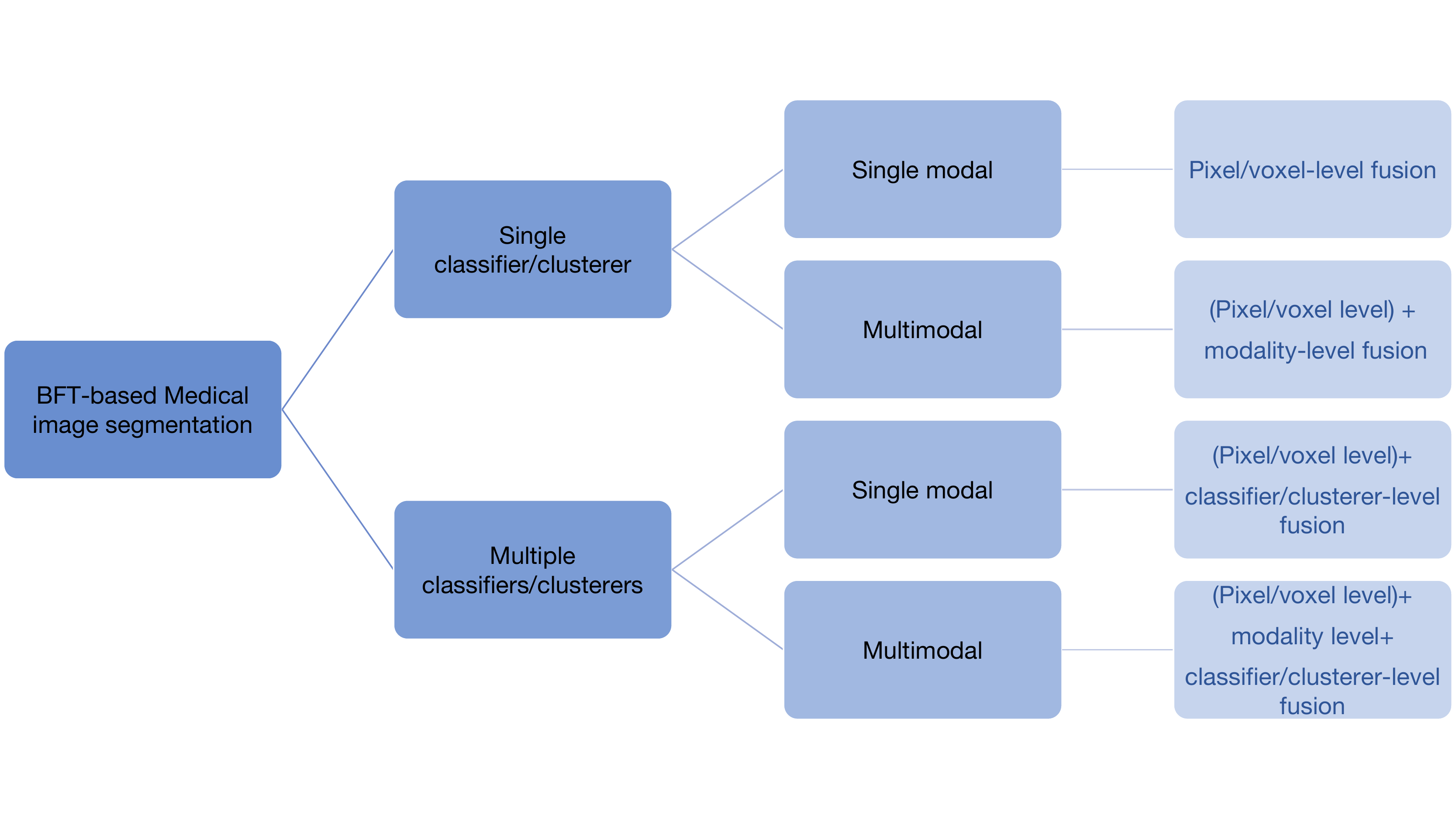}
\caption{Overview of BFT-based medical image segmentation methods.\label{fig:BFT-based-seg}}
\end{figure*}

\begin{figure}
\centering
\includegraphics[width=\textwidth]{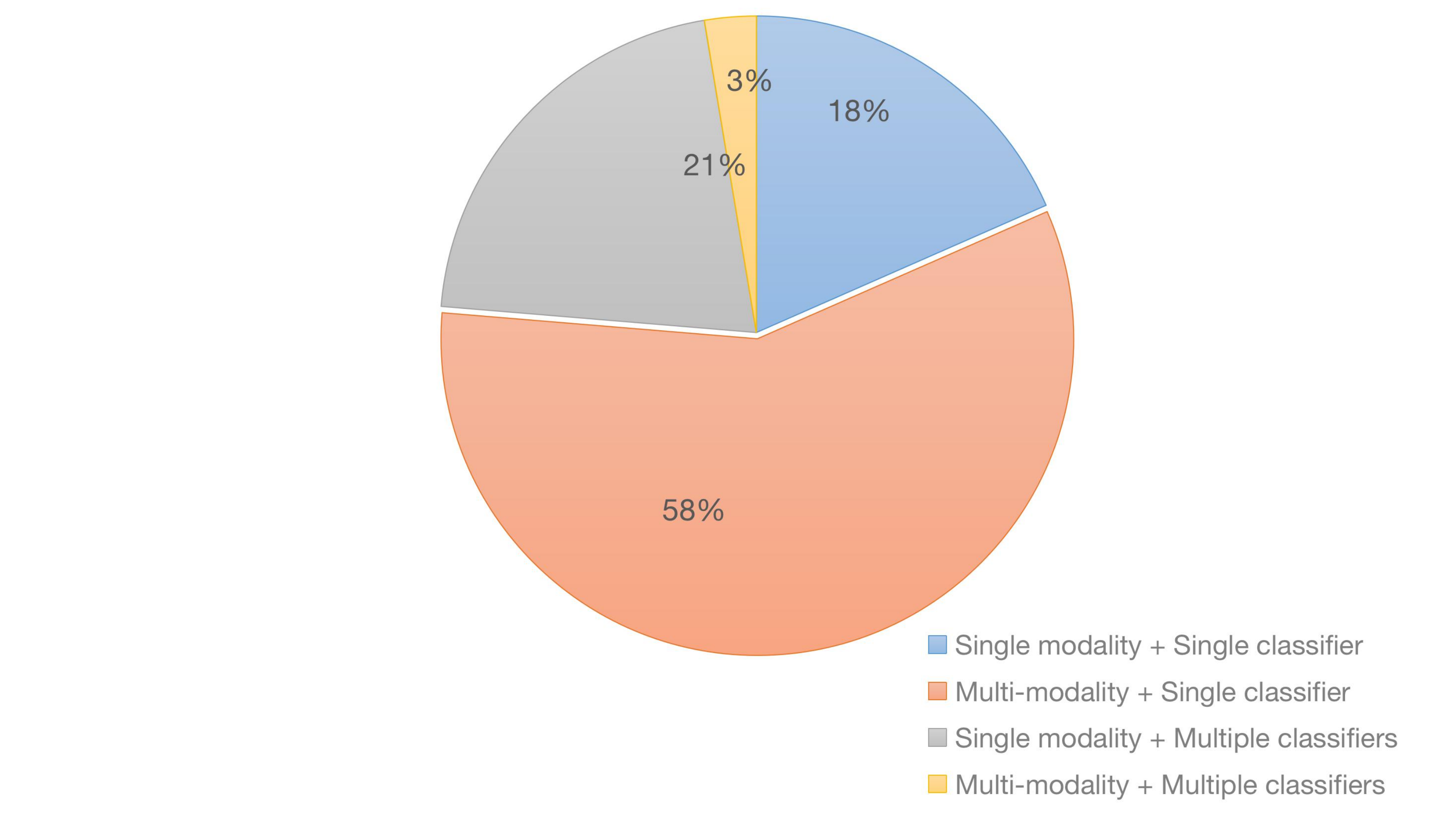}
\caption{The composition of the proportion of modalities and classifiers/clusterers.\label{fig:percentage_2}}
\end{figure}

We presented the common BBA methods to generate mass functions for medical images in Section \ref{sec:mass_finction}. The most interesting way to analyze those methods is to classify them according to the stage at which they fuse mass functions. Figure~ \ref{fig:BFT-based-seg} gives an overview of the BFT-based medical image segmentation methods classified according to the fusion operation they perform.  We first classify methods by the number of classifiers/clusterers, i.e.,~whether the fusion of mass functions is performed at the classifiers/clusterers level. Then we consider the number of input modalities, i.e.,~whether the fusion of mass functions is performed at the modality level. It should be noted that the medical image segmentation with single modal input and single classifier/clusterer can also have a fusion operation, which is performed at the pixel/voxel level. Figure~ \ref{fig:percentage_2} gives the proportions of the BFT-based segmentation methods if multiple classifiers/clusterers or multimodal medical images are used. 76\% of the methods use a single classifier or cluster. Among these methods, 18\% take single modal medical images as input, and 58\% take multimodal medical images as input. The remaining 24\% of the methods use several classifiers or clusterers. Among those methods, 21\% take single modal medical images as input, against 3\% for multimodal medical images. In Sections \ref{subsec: single-clasisifer} and \ref{subsec: multi-clasisifer} we give more details about the fusion operations and discuss their performances.  

\subsection{BFT-based medical image segmentation with a single classifier or clusterer}
\label{subsec: single-clasisifer}
The medical image segmentation methods summarized in this section focus on using a single classifier or clusterer. To simplify the introduction, we classify them into single-modality and multimodal inputs and introduce them in Sections \ref{subsubsec: single} and \ref{subsubsec: multi}, respectively.

\subsubsection{Single-modality evidence fusion}
\label{subsubsec: single}
\begin{figure*}
\includegraphics[width=\textwidth]{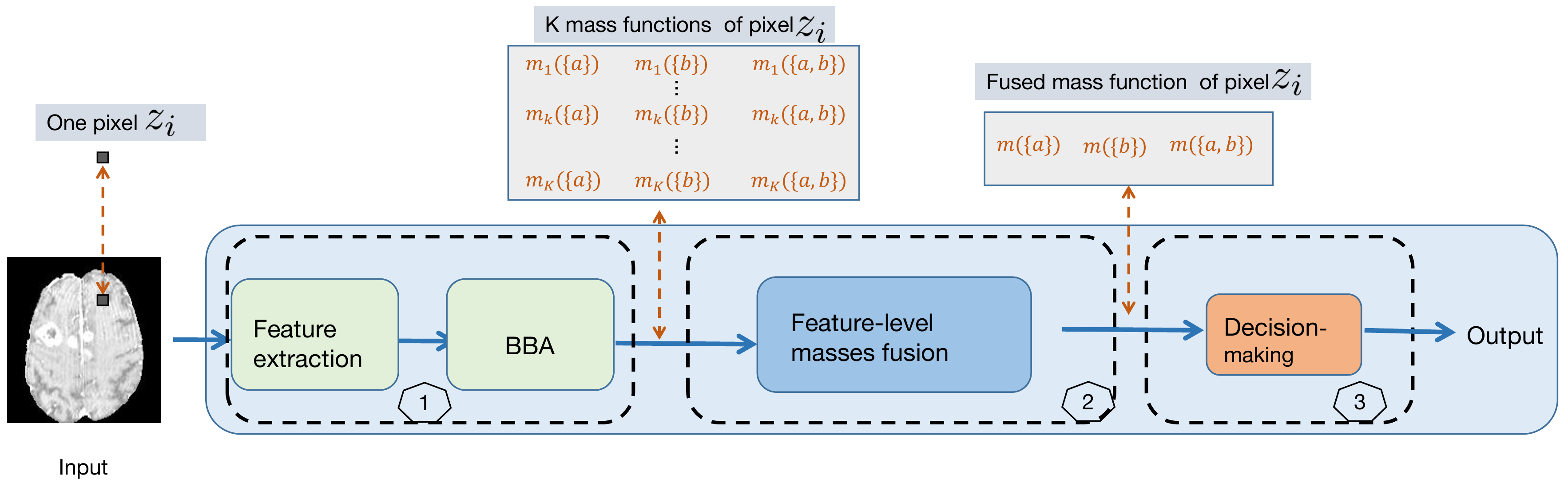}
\caption{Example framework of single modal evidence fusion with single classifier. The segmentation process is composed of three steps: (1) image information is transferred into the feature extraction block, and some BBA methods are used to get feature-level mass functions; (2) Dempster's rule is used to fuse feature-level mass functions; (3) decision-making is made based on the fused mass functions to output a segmented mask. We take only one pixel $z_i$ as an example and show how BFT works on pixel-level evidence fusion under a binary segmentation task to simplify the process. For each pixel $z_i$, we can obtain $K$ mass functions from BBA. After feature-level evidence fusion, a fused mass function is assigned to pixel $z_i$ to represent the degree of belief this pixel belongs to classes a, b and ignorance.\label{fig:fig6}}
\end{figure*}

Figure~\ref{fig:fig6} shows the framework of single-modality evidence fusion with a single classifier (we only take the classifier as an example in the rest of our paper to simplify the explanation). The inputs of the framework are single-modality medical images. The segmentation process comprises three steps: mass calculation (including feature extraction and BBA), feature-level mass fusion, and decision-making. Since decision-making is not the focus of this paper, we will not go into details about it and readers can refer to Ref.~\cite{denoeux2019decision} for a recent review of decision methods based on belief functions.

\begin{itemize}
    \item[(1)]First, the mass calculation step assigns each pixel/voxel {$K$} mass functions based on the evidence of $K$ different BBA methods,  $K$  input features, $K$ nearest neighbors, or $K$ prototype centers. 
    \item[(2)]Dempster's rule is then used for each pixel/voxel to fuse the feature-level mass functions. 
    \item[(3)]Finally, decisions are made to obtain the final segmentation results.
\end{itemize}  
Here, the feature extraction method  is used to generate MVs (corresponding to traditional medical image segmentation methods) or deep features (corresponding to deep learning-based medical image segmentation methods). The feature extraction method could be intensity-based methods such as threshold intensity, probabilistic-based methods such as SVM,  fuzzy set-based methods such as FCM, etc. There are various BBA methods, therefore we introduce them in specific tasks in the following. In general, the methods introduced in this section focus on feature-level evidence fusion. Table \ref{tab:fusion1} shows the segmentation methods with a single modal input and classifier/cluster that focus on feature-level evidence fusion.

\begin{table}
  \centering
  \caption{Summary of BFT-based medical image segmentation methods with single modal inputs and a single classifier/clusterer }
  \scalebox{0.6}{
  \begin{tabular}{llllllll}
 \hline
 \multicolumn{1}{l}{Publications}&
Input image type& Application& BBA method \\

\hline
Suh et al. \cite{SuhKnowledge}&MR images&cardiac segmentation& Shafer's model\\
Gerig et al.\cite{gerig2000exploring}&MR images&brain tissue segmentation&BFOD\\
Vauclin et al. \cite{1509843} &CT&lung and spinal canal segmentation &Shafer's model\\
Vannoorenberghe et al. \cite{vannoorenberghe2006belief} &CT&thoracic segmentation& EKNN \\
Derraz et al. \cite{derraz2013globally} &optical imaging with color&cell segmentation &Appriou's model\\
Derraz et al. \cite{derraz2013image}&optical imaging with color&retinopathy segmentation &Appriou's model \\
Derraz et al. \cite{derraz2013globally}&optical imaging with color& retinopathy segmentation &Appriou's model\\
Lian et al. \cite{lian2017spatial} &FDG-PET/CT&lung tumor segmentation& ECM\\
Lian et al. \cite{lian2017tumor} &FDG-PET&lung tumor segmentation &ECM\\
Huang et al. \cite{huang2021evidential}&PET-CT&lymphoma segmentation&ENN \\
Huang et al. \cite{huang2022lymphoma}&PET-CT&lymphoma segmentation&ENN+RBF\\
\hline
\end{tabular}
}
\label{tab:fusion1}
\end{table}
In~\cite{SuhKnowledge}, Suh et al. developed a knowledge-based endocardial and epicardial boundary detection and segmentation system with cardiac MR image sequences. Pixel and location information were mapped into mass functions by Shafer's model~\cite{shafer1976mathematical} (see Section \ref{subsubsection:lik_methods}). The  mass functions from the two sources  were fused by using Dempster's rule. Experiments were applied to cardiac short-axis images and obtained an excellent success rate ($> 90\%$). However, this work only focused on cardiac boundary detection and did not discuss the details of the heart. In~\cite{1509843}, Vauclin et al. proposed a BFT-based lung and spinal canal segmentation model. The k-means clustering algorithm was first used to perform a pre-segmentation. Then a 3D filter exploits the results of the pre-segmentation to compute the MVs from spatial neighbors using Shafer's model. Segmentation results showed the credal partition permits the reduction of the connection risks between the trachea and the lung when they are very close and between the left and right lungs at the anterior or posterior junctions.

In~\cite{gerig2000exploring}, Gerig et al. presented a method for automatic segmentation and characterization of object changes in time series of 3D MR images. A set of MVs was derived from time series according to brightness changes. BFOD transformation~\cite{safranek1990evidence} was used here to map the obtained features into mass functions. Then the set of evidence was combined by using Dempster's rule. Experiments visually compared with results from alternative segmentation methods revealed an excellent sensitivity and specificity performance in the brain lesion region.
The author also pointed out that better performance could be obtained with multimodal and multiple time-series evidence fusion. Simulation  results showed that about 80\% of the implanted voxels could be detected for most generated lesions.

In~\cite{vannoorenberghe2006belief}, Vannoorenberghe et al. presented a BFT-based thoracic image segmentation method. First, a K-means algorithm performed coarse segmentation on the original CT images. Second, the EKNN rule~\cite{denoeux1995k} was applied by considering spatial information and calculating feature-level mass functions. The authors claimed that using this segmentation scheme leads to a complementary approach combining region segmentation and edge detection. Experimental results showed promising results on 2D and 3D CT images for lung segmentation.

In~\cite{derraz2013globally}, Derraz et al. proposed an active contour
(AC)-based~\cite{chan2000active} global segmentation method for vector-valued image that incorporated both probability and mass functions. All features issued from the vector-valued image were integrated with inside/outside descriptors to drive the segmentation results by maximizing the Maximum-Likelihood distance between foreground and background. Appriou's second model~\cite{appriou200501} was used to calculate the imprecision caused by low contrast and noise between inside and outside descriptors issued from the multiple channels. Then the fast algorithm based on Split Bregman~\cite{goldstein2009split} was used for final segmentation by forming a fast and accurate minimization algorithm for the Total Variation (TV) problem. Experiments were conducted on color biomedical images (eosinophil, lymphocyte, eosinophil, monocyte, and neutrophil cell~\cite{mohamed2012enhanced}) and achieve around 6\% improvements by using F-score on five image groups. In the same year, Derraz et al. proposed a new segmentation method~\cite{derraz2013image} based on Active Contours (AC) for the vector-valued image that incorporates Bhattacharyya's distance~\cite{michailovich2007image}. The only difference is that the authors calculate the probability functions by Bhattacharyya distance instead of the Maximum-Likelihood distance in this paper. The performance of the proposed algorithm was demonstrated on the retinopathy dataset~\cite{quellec2008optimal,niemeijer2009retinopathy}and with an increase of 3\% in F-score compared with the best-performed methods.

In~\cite{lian2017spatial}, Lian et al. introduced a tumor delineation method in fluorodeoxyglucose positron emission tomography (FDG-PET) images by using spatial ECM~\cite{lelandais2014fusion} with adaptive distance metric. The authors proposed the adaptive distance metric to extract the most valuable features, and spatial ECM was used to calculate mass functions. Compared with ECM and spatial ECM, the proposed method showed a 14\% and 10\% increase in Dice score when evaluated on the FDG-PET images of non-small cell lung cancer (NSCLC) patients, showing good performance.

In~\cite{huang2021evidential}, Huang et al. proposed a 3D PET/CT lymphoma segmentation framework with BFT and deep learning. This is the first work that applied BFT with a deep neural network for medical image segmentation. In this paper, the PET and CT images were concatenated as a signal modal input method and transferred into UNet to get high-level semantic features. Then the ENN classifier was used to map high-level semantic features into mass functions by fusing the contribution of $K$ prototypes. Moreover, the segmentation uncertainty was considered in this paper with an uncertainty loss during training. The reported quantitative and qualitative results showed that the proposal outperforms the state-of-the-art methods. Based on the first work, Huang et al. verified the similarity of RBF and ENN in uncertainty quantification and merged them with the deep neural network (UNet) for lymphoma segmentation~\cite{huang2022lymphoma}. The segmentation performance confirmed that RBF is an alternative approach of ENN~\cite{denoeux19d} to act as an evidential classifier and showed that the proposal outperforms the baseline method UNet and the state-of-the-art both in accuracy and reliability. 

Before 2020, the BFT-based medical image segmentation methods with single modal medical used low-level image features, e.g.,~grayscale and shape features, to generate mass functions, which limits the segmentation accuracy. Moreover, None of them discussed the reliability of the segmentation results. Huang et al. first merge BFT with a deep segmentation model (UNet) and learn the representation of mass functions with some learning algorithms~\cite{huang2021evidential}. Based on this, Huang et al. further discuss the relationship between segmentation accuracy and reliability in~\cite{huang2022lymphoma}, which takes a new direction to study reliable medical image segmentation methods and bridge the gap between experimental results and clinical application.

 \subsubsection{Multimodal evidence fusion}
\label{subsubsec: multi}
Single-modality medical images often do not contain enough information to present the information about the disease and are often tainted with uncertainty. In addition to feature-level evidence fusion, the fusion of multimodal evidence is also important to achieve accurate and reliable medical image segmentation performance. Approaches for modality-level evidence fusion can be summarized into three main categories according to the way they calculate the evidence: probabilistic-based fusion, fuzzy set-based fusion, and BFT-based fusion. The development of convolution neural networks (CNNs) further contributes to the probabilistic-based fusion methods~\cite{zhou2019review}, which can be summarized into image-level fusion (e.g.,~data concatenation~\cite{peiris2021volumetric}), feature-level fusion (e.g.,~attention mechanism concatenation~\cite{zhou2020fusion,zhou2022tri}), and decision-level fusion (e.g.,~model ensembles~\cite{kamnitsas2017ensembles}). However, none of those methods considers the conflict of source evidence, i.e., the modality-level uncertainty is not well studied, which limits the reliability and explainability of the performance.

\begin{figure*}
\includegraphics[width=\textwidth]{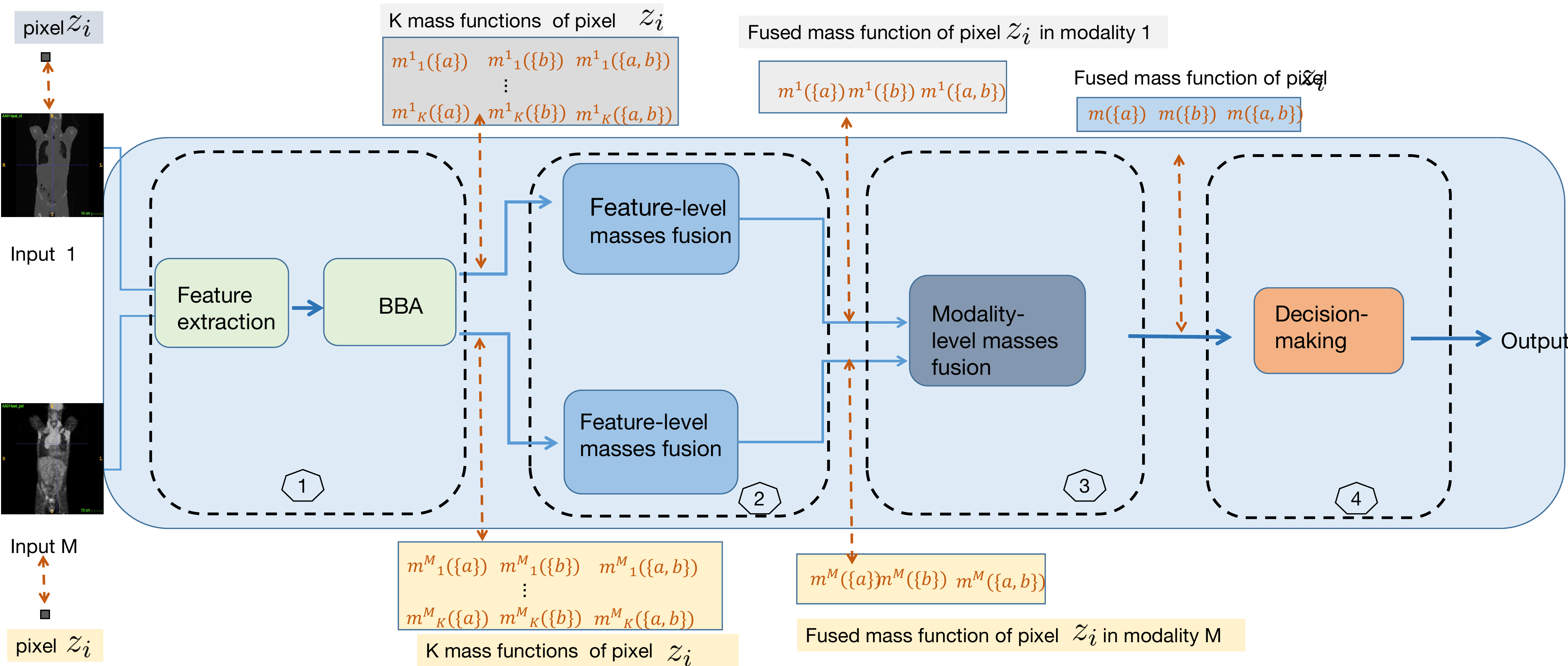}
\caption{Example framework of multimodal evidence fusion with a single classifier. The segmentation process is composed of four steps: (1) for each modal input, image features are fed into the classifier, and some BBA methods are used to get feature-level mass functions; (2) inside each modal, the calculated feature-level mass functions are fused by using Dempster's rule to generate modality-level mass functions; (3) between the modalities, the calculated modality-level mass functions are fused by using Dempster's rule again; (4) decision-making is made based on the fused mass functions to output a segmented mask. To simplify the introduction, we show the segmentation example of the same located pixels $z_i^1$ and $z_i^M$ obtained from modal 1 and $M$. The same located pixels from different modalities are transferred separately into one classifier, and some BBA is used to get pixel-level mass functions. Similar to Figure \ref{fig:fig6}, for each pixel, assume we can obtain $K$ mass functions. After feature-level and modality-level evidence fusion, a fused mass function is assigned to the pixel $z_i$ to represent the degree of belief this pixel belongs to classes a, b and ignorance.\label{fig:fig7}}
\end{figure*}

This section focuses on the BFT-based segmentation methods with modality-level evidence fusion. Figure~\ref{fig:fig7} shows an example framework of multimodal evidence fusion with a single classifier. We separate the segmentation process into four steps: masses calculation, feature-level evidence fusion (alternative), modality-level evidence fusion, and decision-making. Compared with single-modality evidence fusion reviewed in Section \ref{subsubsec:single-modality}, multimodal evidence fusion focuses here not only on feature-level but also on modality-level evidence fusion (It should be noted that feature-level evidence fusion is not necessary in this case). Multimodal evidence fusion is the most popular application for BFT in the medical image segmentation domain. Therefore we classify those methods according to their input modal for better analysis.  Table \ref{tab:fusion2} summarizes the segmentation methods with multimodal inputs and a single classifier/cluster with the main focus on modality-level evidence fusion.

\begin{table}
  \centering
  \caption{Summary of BFT-based medical image segmentation methods with multimodal inputs and single classifier/clusterer }
  \scalebox{0.6}{
  \begin{tabular}{llllllll}
 \hline
 \multicolumn{1}{l}{Publications}&Input image type& Application & BBA method \\

\hline
Vannoorenberghe et al.
\cite{vannoorenberghe1999dempster}&optical imaging with color&skin cancer segmentation& GD-based model\\
Bloch \cite{bloch1996some}&multimodal MR images&brain tissue segmentation& prior knowledge \\
Taleb-Ahmed \cite{taleb2002information}&multimodal MR images&vertebrae segmenta& threshold+contour distance \\
Chaabane et al. \cite{chaabane2009relevance}&optical imaging with color&cell lesion segmentation & Appriou's model \\
Zhu et al. \cite{zhu2002automatic} &multimodal MR images& brain tissue segmentation &Zhu's model \\
Bloch \cite{bloch2008defining}&multimodal MR images &brain tissue segmentation& prior knowledge \\
Chaabane et al. \cite{chaabane2011new}  &optical imaging with color&cell lesion segmentation&Aprrious's method\\
Ghasemi et al. \cite{ghasemi2012brain}&multimodal MR images& brain tissue segmentation  &RMV\\
Harrabi et al. \cite{harrabi2012color} & optical imaging with color& breast cancer segmentation &GD-based model\\
Lelandais et al. \cite{lelandais2012segmentation} &PET &biological target tumor segmentation &ECM\\
Wang et al. \cite{Rui2013Lesion} &multimodal MR images &cerebral infraction segmentation &Zhu's model \\
Ghasemi et al. \cite{ghasemi2013novel} &multimodal MR images& brain tumor segmentation & RMV\\
Lelandais et al. \cite{lelandais2014fusion}& multi-tracer PET&biological target tumor segmentation& ECM\\
Makni et al. \cite{makni2014introducing}&multi-parametric MR image & prostate zonal anatomy &ECM \\
Derraz et al. \cite{derraz2015joint}  &PET/CT &lung tumor segmentation  & Appriou's model\\
Trabelsi et al. \cite{trabelsi2015skin} &optical imaging with color &skin lesion segmentation&None \\
Xiao et al. \cite{xiao2017vascular} &multimodal MR images& vascular segmentation &GD-based model\\
Lian et al. \cite{lian2017accurate} &FDG-PET/CT& lung tumor segmentation&ECM \\
Lian et al. \cite{lian2018unsupervised} &FDG-PET/CT &lung cancer segmentation  &ECM\\
Tavakoli et al. \cite{tavakoli2018brain} &multimodal MR images &brain tissue segmentation &RMV\\
Lima et al. \cite{lima2019modified} &multimodal MR images& brain tissue segmentation&RMV \\
Lian et al. \cite{lian2018joint} &PET/CT& lung tumor segmentation &ECM\\
\hline
\end{tabular}
}
\label{tab:fusion2}
\end{table}
\paragraph*{Fusion of multimodal MR images}
 
In~\cite{bloch1996some}, Bloch first proposed a BFT-based dual-echo MR pathological brains tissue segmentation model with uncertainty and imprecision quantification. The author assigned mass functions based on a reasoning approach that uses gray-level histograms provided by each image to choose focal elements. After defining mass functions, Dempster's rule combined the mass from dual-echo MR images for each pixel. Based on the first work with BFT, in~\cite{bloch2008defining}, Bloch proposed to use fuzzy mathematical morphology~\cite{bloch1995fuzzy}, i.e.,~erosion and dilation, to generate mass functions by introducing imprecision in the probability functions and
estimating compound hypotheses. Then Dempster's rule is used to fuse mass functions from multimodal images. It should be noted that in this paper, the strong assumption is made that it is possible to represent imprecision by a fuzzy set, also called the structuring element. Application examples on dual-echo MR image fusion showed that the fuzzy mathematical morphology operations could represent either spatial domain imprecision or feature space imprecision (i.e., gray levels features). The visualized brain tissue segmentation results show the robustness of the proposed method.

As we mentioned in Section \ref{subsub:MVs}, Zhu et al. proposed modeling mass functions in BFT using FCM and spatial neighborhood information for image segmentation. The visualized segmentation results on MR brain images showed that the fusion-based segmented regions are relatively homogeneous, enabling accurate measurement of brain tissue volumes compared with single modal input MR image input.

In~\cite{ghasemi2012brain}, Ghasemi et al. presented a brain tissue segmentation approach based on FCM and BFT. The authors used the FCM to model two different input features: pixel intensity and spatial information, as membership values (MVs). Then for each pixel, the RMV transformation~\cite{ghasemi2012brain} was used to map MVs into mass functions. Last, the authors used Dempster's rule to fuse intensity-based and spatial-based mass functions to get final segmentation results. Compared with FCM, the authors reported an increase in Dice and accuracy. As an extension of~\cite{ghasemi2012brain}, Ghasemi et al. proposed an unsupervised brain tumor segmentation method that modeled pixel intensity and
spatial information into mass functions with RMV transformation and fused the two mass functions with Dempster's rule in~\cite{ghasemi2013novel}. 

In~\cite{Rui2013Lesion}, Wang et al. proposed a lesion segmentation method for infarction and cytotoxic brain edema. The authors used a method similar to Zhu's model to define simple and double hypotheses. FCM~\cite{bezdek1984fcm} was used first to construct the mass functions of simple hypotheses $\{a\}$ and $\{ b\}$. Then masses were assigned to double hypotheses as $m(\{a, b\})=\frac{1}{4} \times \frac{min(m(\{a\}),m(\{b\}))}{m(\{a\})+m(\{b\})}$. Finally, the authors used Dempster's rule for modality-level evidence fusion. The results showed that infarction and cytotoxic brain edema around the infarction could be well segmented by final segmentation.

In~\cite{makni2014introducing}, Makni et al. introduced a  multi-parametric MR image segmentation model by using spatial neighborhood in ECM for prostate zonal anatomy. The authors  extended ECM with neighboring voxels information to map multi-parametric MR images into mass functions. Then they used prior knowledge related to defects in the acquisition system to reduce uncertainty and imprecision. Finally, the authors used Dempster's rule to fuse the mass functions from the multi-parametric MR images. The method achieved good performance on prostate multi-parametric MR image segmentation.

In~\cite{tavakoli2018brain}, Tavakoli et al. proposed a segmentation method based on the evidential fusion of different modalities (T1, T2, and Proton density (PD)) for brain tissue. The authors used FCM to get MVs and used the RMV transformation~\cite{ghasemi2012brain} to transform the clustering MVs into mass functions. The authors first formed the belief structure for each modal image and used 
Dempster's rule to fuse the three modalities' mass functions of T1, T2, and PD. Compared with FCM, this method achieved a 5\% improvement in the Dice score. Based on Tavakoli's method~\cite{tavakoli2018brain}, Lima et al. proposed a modified brain tissue segmentation method in~\cite{lima2019modified}. The authors tested their method with four modality inputs: T1, T2, PD, and Flair. The reported results outperformed both the baseline method FCM and Tavakoli's method with four-modality evidence fusion. 

In~\cite{xiao2017vascular}, Xiao et al. proposed an automatic vascular segmentation algorithm, which combines the grayscale and shape features of blood vessels and extracts 3D vascular structures from the head phase-contrast MR angiography dataset. First, grayscale and shape features are mapped into mass functions by using the GD-based method. Second, a new reconstructed vascular image was established according to the fusion of vascular grayscale and shape features by Dempster's rule. Third, a segmentation ratio coefficient was proposed to control the segmentation result according to the noise distribution characteristic. Experiment results showed that vascular structures could be detected when both grayscale and shape features are robust. Compared with traditional grayscale feature-based or shape feature-based methods, the proposal showed better performance in segmentation accuracy with the decreased over-segmentation and under-segmentation ratios by fusing two sources of information.

Since Bloch's early work fully demonstrated the advantages of BFT in modeling uncertain and imprecision, introducing partial or global ignorance, and fusing conflicting evidence in a multimodal MR images segmentation task, researchers in this domain have gone further to explore the advantages of BFT in multi MR image fusion. Among these research works, FCM is the most popular clustering algorithm to map input information into MVs. Ration MV transformation or Zhu's model is usually used to generate mass functions from MVs. Apart from these two-step methods, the GD-based model can also be used to generate mass functions directly. 

\paragraph*{Fusion of RGB channels}

In~\cite{vannoorenberghe1999dempster}, Vannoorenberghe et al. pointed out that taking the R, G, and B channels as three independent information sources can be limited and nonoptimal for medical image tasks and proposed a color image segmentation method based on BFT. They calculated the degree of evidence by mapping R, G, and B channel intensity into mass functions using the Gaussian distribution information (similar to GD-based model~\cite{chen2012manifold}) with an additional discounting operation). Then three pieces of evidence were fused with Dempster's rule. The proposed segmentation method was applied to biomedical images to detect skin cancer (melanoma). Experiments showed a significant part of the lesion is correctly extracted from the safe skin. The segmentation performance is limited by feature representation, e.g., some regions correspond to pixels that cannot be classified as either the safe skin or the lesion because only the pixel-level feature is insufficient for hard-example segmentation.

In~\cite{chaabane2009relevance} Chaabane et al. proposed a color medical image segmentation method based on fusion operation. Compared to~\cite{vannoorenberghe1999dempster}, the authors first modeled  probabilities for each region by a Gaussian distribution, then used Appriou's second model (see Section \ref{eq:appriou2}) to map probability into mass functions. Dempster's rule then combined the evidence from the three color channels. Compared with single-channel segmentation results, the fused results achieved a 10\% increase in segmentation sensitivity. 

Different from the methods described in~\cite{vannoorenberghe1999dempster} and \cite{chaabane2009relevance} that decompose color images into R, G, B channels, Harrabi et al.~\cite{harrabi2012color} presented a color image segmentation method that represents the color image with 6 color spaces (RGB, HSV, YIQ, XYZ, LAB, LUV). The segmentation operation is based on multi-level thresholding and evidence fusion techniques. First, the authors identified the most significant peaks of the histogram by multi-level thresholding with the two-stage Otsu optimization approach. Second, the GD-based model was used to calculate the mass functions for each color space. Then the authors used
Dempster's rule to combine six sources of information. Compared with single color spaces, such as RGB and HSV, the combined result taking into account six color spaces, has a
significant increase in segmentation sensitivity, for example, an increase of 4\% and
7\% as compared to RGB and HSV, respectively.

In~\cite{trabelsi2015skin}, Trabelsi et al. applied BFT in optical imaging with color to improve skin lesion segmentation performance. The authors decomposed color images into R, G, and B channels and applied the FCM method on each channel to get probability functions for pixel $x$ in each color space. In this paper, the authors take the probability functions as mass functions and calculate the orthogonal sum of the probability functions from the three-channel images. Even though experiments showed about 10\% improvements in
segmentation accuracy compared with single-channel results, this work does not harness the full power of BFT as it only considers Bayesian mass functions.

In general, the BFT-based RGB medical image segmentation approaches are used to generate mass functions from possibility distributions, e.g.,~Gaussian distribution and Possibility C-means distribution, and fuse them by Dempster's rule. Though the authors claimed they could get better performance compared with single color input, the segmentation performance is limited by features because gray-scale and intensity features are not robust and efficient in representing image information. Further research could take both feature extraction and uncertainty quantification into consideration, e.g., a deep feature extraction model with an evidential classifier, to improve the performance.

\paragraph*{Fusion of PET/CT}

In~\cite{lelandais2012segmentation}, Lelandais et al. proposed a BFT-based multi-trace PET images segmentation method to segment biological target volumes. First, the authors used a modified FCM algorithm with the discounting algorithm to determine mass functions. The modification integrated a disjunctive combination of neighboring voxels inside the iterative process. Second, the operation of reduction of imperfect data was conducted by fusing neighboring voxels using Dempster's rule. Based on this first work, Lelandais
et al. proposed an ECM-based fusion model~\cite{lelandais2014fusion} for biological target volume segmentation with multi-tracer PET images. The segmentation method introduced in this paper is similar to the one introduced in~\cite{makni2014introducing} with a different application.

In~\cite{derraz2015joint}, Derraz et al. proposed a multimodal tumor segmentation framework for PET and CT inputs. Different from Lelandais et al.'s work that uses ECM or optimized ECM to generate mass functions directly, the authors construct mass functions in two steps. They first proposed a NonLocal Active Contours (NLAC) based variational segmentation framework to get probability results. Then, similar to the authors' previous
work~\cite{derraz2013globally,derraz2013image}, they used Appriou's second model~\cite{appriou200501} to map MVs into mass functions. Then Dempster's rule was used to fuse the mass functions from PET and CT modalities. The framework was evaluated on a lung tumor segmentation task. Compared with the state-of-the-art methods, this framework yielded the highest Dice score for tumor segmentation. 

Based on Lelandais et al.'s work, Lian et al. proposed a tumor segmentation method~\cite{lian2017accurate} using Spatial ECM~\cite{lelandais2014fusion} with Adaptive Distance Metric~\cite{lian2017spatial} in FDG-PET images with the guidance of CT images. Based on the first work, Lian et al. proposed a co-segmentation method of lung tumor
segmentation~\cite{lian2018joint,lian2018unsupervised}. They took PET and CT as independent inputs and use ECM to generate mass functions. At the same time, an adaptive distance metric was used to quantify clustering distortions and spatial relationships during the evidential clustering procedure. Then Dempster's rule was used to fuse mass functions from PET and CT modalities. The quantitative and qualitative evaluation results showed superior performance compared with single modal segmentation results with an increase of 1\% and 58\% in PET and CT in Dice scores, respectively.

ECM is the most common BBA method to generate mass functions for PET/CT medical image segmentation approaches. Similar to the BFT-based RGB medical image segmentation methods, the segmentation performance here is limited by feature extraction methods. Further research could build on recent advancements in deep feature representation and combine ECM with deep neural networks to learn mass feature representation. A good example of neural-network approach to evidential clustering can be found in~\cite{denoeux2021nn}.

\subsection{BFT-based medical image segmentation with several classifiers or clusterers}
\label{subsec: multi-clasisifer}
It is common practice that two or more physicians cooperate  for disease diagnosis, which can minimize the impact of physicians' misjudgments. Similarly, combining the results from multiple decision mechanisms as well as addressing the conflicts is critical to achieving a more reliable diagnosis. This section introduces the BFT-based medical image segmentation methods with several classifiers or clusterers. We follow the same approach as in Section \ref{subsec: single-clasisifer} and separate the methods into single-modality and multimodal evidence fusion reviewed, respectively, in Sections \ref{subsubsec:single-modality} and \ref{subsubsec:multi-modality}.
 
 \subsubsection{Single-modality evidence fusion}
\label{subsubsec:single-modality}
\begin{figure*}
\includegraphics[width=\textwidth]{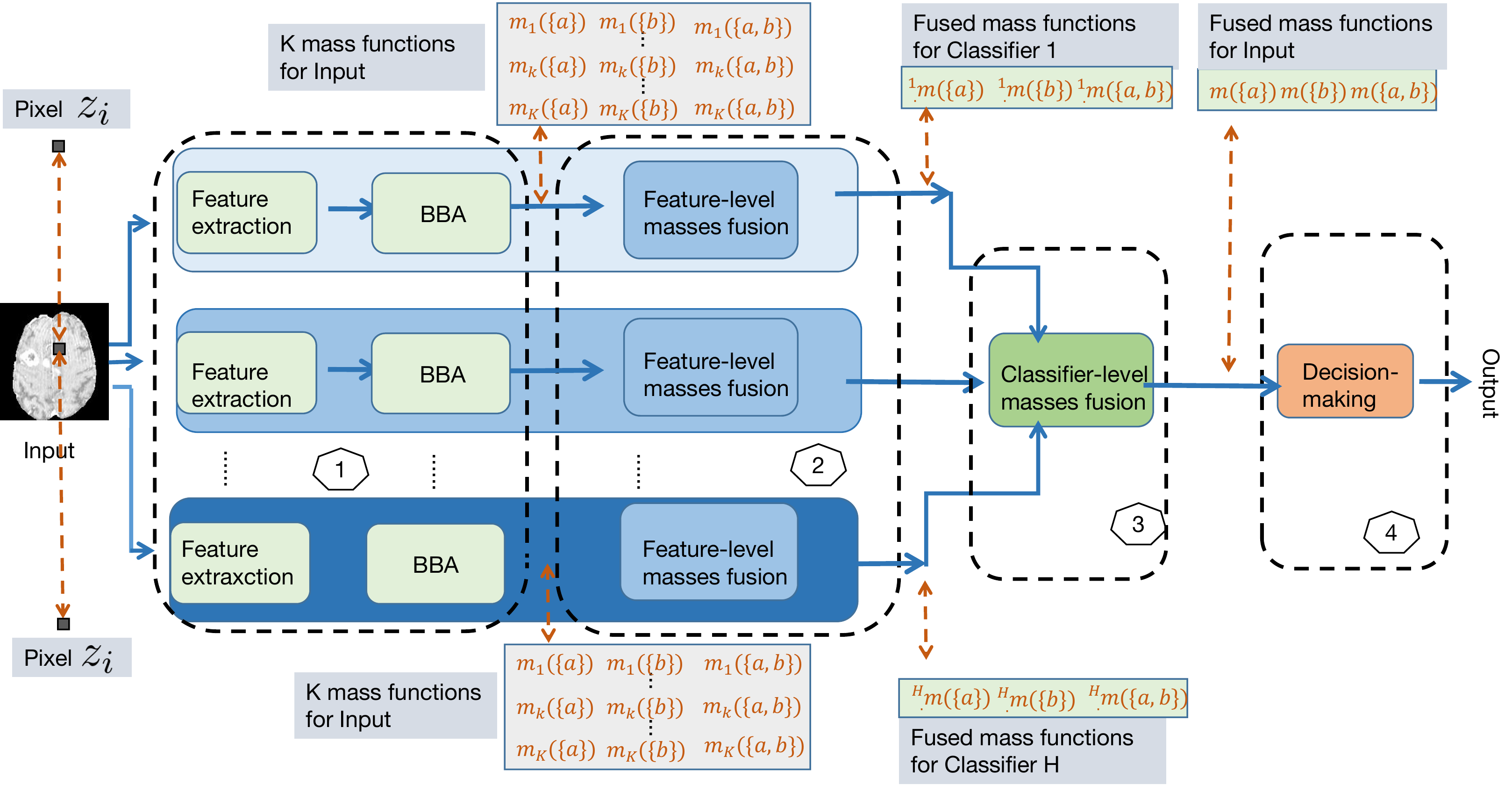}
\caption{Example framework of single modal evidence fusion with several classifiers. The segmentation process is composed of four steps: (1) for each pixel, image features are transferred into the different classifiers, and some BBA methods are used to get feature-level mass functions; (2) inside each classifier, the calculated feature-level mass functions are fused by using Dempster's rule to generate classifier-level mass functions; (3) between the classifiers, the calculated classifier-level mass functions are fused by using Dempster's rule again; (4) decision-making is made based on the final fused mass functions and output a segmented mask. We take pixel $z_i$ as an example. Similar to Figure~ \ref{fig:fig6}, we assume for each pixel, we can obtain $K$ mass functions by one classifier. After feature-level evidence fusion, for each pixel $z_i$, we can get $H$ mass functions corresponding to $H$ classifiers. Then we fuse the $H$ mass functions and assign a fused mass function to the pixel $z_i$, representing the degree of belief this pixel belongs to classes a, b and ignorance.\label{fig:fig8}}
\end{figure*}

\begin{table}
  \centering
  \caption{Summary of BFT-based medical image segmentation methods with single modal inputs and several classifiers/clusterers }
  \scalebox{0.6}{
  \begin{tabular}{llllllll}
 \hline
 \multicolumn{1}{l}{Publications} &Input image type& Application&BBA method \\
\hline
Capelle et al. \cite{capelle2002segmentation}&MR images &brain tumor segmentation&EKNN\\
Capelle et al. \cite{capelle2004evidential}&MR images& brain tumor segmentation&EKNN+Shafer's model + Appriou's model\\
Barhoumi et al. \cite{barhoumi2007collaborative} &optical imaging with color &skin lesion malignancy tracking& None \\
Guan et al. \cite{guan2011study}&MR images& brain tissue segmentation&Zhu's model\\
Ketout et al. \cite{ketout2012improved} &optical imaging with color& endocardial contour detection& Threshold \\
Wen et al. \cite{wen2018improved} &MR images &brain tissue segmentation &Zhu's model+GD-based model\\
George et al. \cite{george2020breast} &optical imaging with color &breast cancer segmentation&Discounting \\
Huang et al. \cite{huang2021evidential} &PET-CT &lymphoma segmentation &ENN\\

\hline
\end{tabular}
}
\label{tab:fusion3}
\end{table}
Compared with the methods presented in Section \ref{subsubsec: single}, the methods introduced in this section focus on feature-level and classifier-level evidence fusion, which aims to minimize the impact of misjudgments caused by a single model's inner shortcomings. Figure \ref{fig:fig8} shows an example of a medical image segmentation framework with single-modality input and several classifiers. We separate the segmentation process into four steps: mass calculation, feature-level evidence fusion (optional), classifier-level evidence fusion, and decision-making. Table \ref{tab:fusion3} summarizes the segmentation methods with single-modality inputs and several single classifiers/clusterers with the main focus on classifier/clusterer level evidence fusion.

In~\cite{capelle2002segmentation}, Capelle et al. proposed a segmentation scheme for MR images based on BFT. The authors used the Evidential K-NN rule~\cite{denoeux1995k} to map image features into mass functions. Then, they applied the evidential fusion process to classify the voxels. Based on this first work, Capelle et al. later proposed an evidential segmentation scheme of multimodal MR images for brain tumor detection in~\cite{capelle2004evidential}. This work focused on analyzing different evidential modeling techniques and on the influence of the introduction of spatial information to find the most effective brain tumor segmentation method. Three different BFT-based models: the distance-based BFT model (EKNN)~\cite{denoeux2000neural}, the likelihood function-based BFT method (Shafer's model~\cite{shafer1976mathematical}), and Appriou's first model~\cite{appriou1999multisensor} were used to model information, and Dempster's rule was used to combine the three mass functions. This study concluded that neighborhood information increases the evidence of class membership for each voxel, thus making the final decision more reliable. Experimental results showed better segmentation performance compared with the state-of-the-art methods when the paper was published.

In~\cite{taleb2002information}, Taleb-Ahmed proposed a segmentation method for MR sequences of vertebrae in the form of images of their multiple slices with BFT. The authors used three different classifiers to calculate three kinds of mass functions. Firstly, the authors used gray-level intensity and standard deviation information to calculate two pixel-level mass functions with two fixed thresholds. Then the distance between two matching contours of consecutive slices ($P$ and $Q$) was used to calculate contour-level mass functions as follows:
\begin{subequations}
\begin{equation}
m(\{S_{PQ}\})=\begin{cases}
 1-e^{-\eta \left |   d(P_{i},Q_{i})-\beta \right |   } & \text{if } d(P_{i},Q_{i}) \in [\rho,\beta],\\ 
 0 & \text{otherwise}
 \end{cases}
 \end{equation}
 \begin{equation}
 m(\overline{\{S_{PQ}\}})= 
 \begin{cases}
 1-e^{-\eta \left |  d(P_{i},Q_{i})-\beta \right |   }& \text{if } d(P_{i},Q_{i}) \in (\beta,+\infty),\\ 
 0 & \text{otherwise}
 \end{cases}
 \end{equation}
  \begin{equation}
 m(\Omega)=e^{-\eta \left |   d(P_{i},Q_{i})-\beta \right |  },
\end{equation} 
 \label{eq:36}
\end{subequations}
where $P_{i}$ and $Q_{i}$ are two matching points of the slices $P$ and $Q$, $d(P_{i},Q_{i})$ is the corresponding distance; $\Omega=\{S_{PQ},\overline{S_{PQ}}\}$, $S_{PQ}$ means that  points $P_{i}$ and $Q_{i}$  both belong to cortical osseous. Parameter $\beta$ represents the tolerance that the expert associates with the value $d(P_{i}, Q_{i})$, $\rho$ is the inter-slice distance and $\eta$ makes it possible to tolerate a greater inaccuracy in the geometrical resemblance of two consecutive contours. Dempster's rule was then used to combine the three mass functions for final segmentation.

In~\cite{barhoumi2007collaborative}, Barhoumi et al. introduced a new collaborative computer-aided diagnosis system for skin malignancy tracking. First, two diagnosis opinions were produced by perceptron neural network classification and content-based image retrieval (CBIR) schemes. Then Dempster's rule was used to fuse the two opinions to achieve a final malignancy segmentation. Simulation results showed that this BFT-based combination could generate accurate diagnosis results. In this work, the frame of discernment is composed of two elements, and the mass functions are Bayesian.

In~\cite{guan2011study}, Guan et al. proposed a human brain tissue segmentation method with BFT. The authors first used Markov random forest (MRF)~\cite{li2009markov} for spatial information segmentation and then a two-dimensional histogram (TDH) method of fuzzy clustering~\cite{duan2008multi} to get a vague segmentation. Then a redundancy image was generated, representing the difference between the MRF and TDH methods, and Zhu's model~\cite{zhu2002automatic} was used to calculate mass functions. Finally, Dempster's rule fused the two segmentation results and the generated redundancy image to handle controversial pixels. The visual segmentation results showed that this method has higher segmentation accuracy compared with the state-of-the-art.

As discussed in Section \ref{subsec:fusion}, Dempster's rule becomes numerically unstable when combining highly conflicting mass functions. In this case, the fused results can be unreliable, as a small changes in mass functions can result in sharp changes of the fusion results. Researchers in the medical domain have also recognized this limitation. In~\cite{ketout2012improved}, Ketout et al. proposed a modified mass computation method to address this limitation and applied their proposal to endocardial contour detection. First, the outputs of each active contour model (ACM)~\cite{kass1988snakes} were represented as mass functions. Second, a threshold was proposed to check if the evidence $m$ conflicts with others. If there was conflict, a modifying operation was used on the conflicting evidence. Finally, the results of edge set-based segmentation~\cite{li2005level} and region set-based segmentation~\cite{chan2001active} were fused by using the ``improved BFT''~\cite{xin2005improved} to get a more accurate contour of the left ventricle. Experimental results showed that the fused contour is closer to the ground truth than the contour from the edge or region. False detection of the two contours was suppressed in the resulting image by rejecting the conflicting events by the fusion algorithm. Meanwhile, the proposed method could detect the edges of the endocardial borders even in low-contrast regions.

In~\cite{wen2018improved}, Wen et al. proposed an improved MR image segmentation method based on FCM and BFT. First, the authors fused two modality images $A$ and $B$ with  function $F(x,y) = w_1  g_{A}(x, y)+w_2  g_{B}(x, y)$, where $x$ and $y$ are image pixels and $w_1$ and $w_2$ are weighs used to adjust the influence of different images on the final fusion result and verifying $w_1 + w_2=1$. Second, the authors calculated the MV by FCM and calculated the mass functions of simple and double hypotheses by Zhu's model~\cite{zhu2002automatic} (see Section \ref{subsub:MVs}) without the neighboring pixel information. Third, the authors generated another mass functions by weighting those of its neighboring pixels with the GD-based model and used Zhu's model again to construct simple and double hypotheses mass functions. Finally, the authors used Dempster's rule to complete the fusion of the two mass functions to achieve the final segmentation. Compared with the FCM-based method, the proposed method can better decrease the conflict in multiple sources to achieve easy convergence and significant improvement by using Dempster's rule for classifier-level evidence fusion.

Besides, with the development of CNNs, the research community used Dempster's rule for the fusion of multiple CNN classifiers. In~\cite{george2020breast}, George et al. proposed a breast cancer detection system using transfer learning and BFT. This first work first applied BFT in multiple evidence fusion with deep learning. High-level features were extracted using a convolutional neural network such as ResNet-18, ResNet-50, ResNet-101, GoogleNet, and AlexNet. A patch-based feature extraction strategy was used to avoid wrong segmentation of the boundaries and provide features with good discriminative power for classification. The extracted features were classified into benign and malignant classes using a support vector machine (SVM). A discounting operation was applied to transfer probability-based segmentation maps into mass functions. The discounted outputs from the different CNN-SVM frameworks were then combined using Dempster's rule. This work takes advantage of deep learning and BFT and has achieved good performance. Compared with majority voting-based fusion methods, BFT-based fusion showed superior segmentation accuracy. Compared with a single classifier, such as ResNet-101, the fused framework achieved an increase of 1\%, 0.5\%, 3\%, and 2\% for, respectively, sensitivity, specificity, and AUC. Also, the authors compared their results with the state-of-the-art method and achieved comparable segmentation accuracy.

Apart from using BFT to combine the discounted probabilities from the CNN classifiers~\cite{george2020breast}, another solution is to construct a deep evidential segmentation framework directly. In~\cite{huang2021belief}, Huang et al. proposed a BFT-based semi-supervised learning framework (EUNet) for brain tumor segmentation. This work applied BFT in a deep neural network to quantify segmentation uncertainty directly. During training, two kinds of evidence were obtained: the segmentation probability functions and mass functions generated by UNet and EUNet, respectively. Dempster's rule was then used to fuse the two pieces of evidence. Experimental results showed that the proposal has better performance than state-of-the-art methods. It achieved around 10\%, 4\%, and 2\% increase in Dice score, Positive predictive value, and sensitivity, respectively, compared with the baseline (UNet). Moreover, the authors showed how BFT exceeds probability-based methods in addressing uncertainty boundary problems. This is the first work that embeds BFT into CNN and achieves an end-to-end deep evidential segmentation model.

The approaches introduced in this section use several classifiers or clusterers to generate different mass functions  and fuse them by Dempster's rule. Among those approaches, George et al.~\cite{george2020breast} first applied Dempster's rule to combine the discounted probabilities from different deep segmentation models. Huang et al.~\cite{huang2021belief} merged ENN with UNet to construct an end-to-end segmentation model and fuse two kinds of evidence by Dempster's rule. Compared to George et al.'s approach~\cite{george2020breast}, Huang et al.'s approach~\cite{huang2021belief} can generate mass functions directly from a deep segmentation model, which is more promising.
 \subsubsection{Multimodal evidence fusion}
\label{subsubsec:multi-modality}
\begin{figure*}

\includegraphics[width=\textwidth]{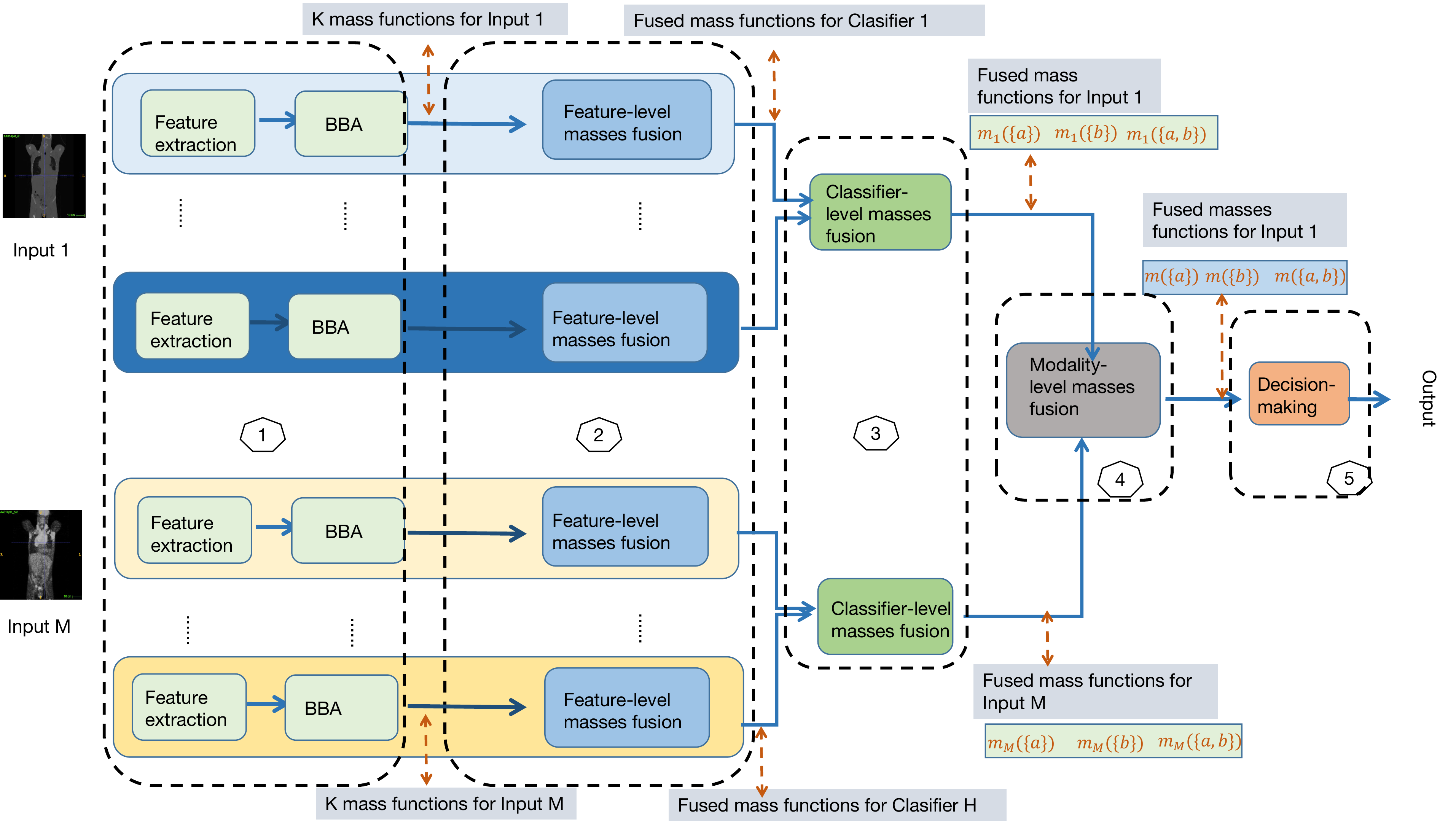}
\caption{Example framework of multimodal evidence fusion with several classifiers. The segmentation process is composed of five steps: (1) for each modality input, image features are fed into different classifiers, and some BBA methods are used to get feature-level mass functions; (2) inside each classifier, the feature-level mass functions are fused by Dempster's rule to get classifier-level mass functions; (3) inside each classifier, the calculated classifier-level mass functions are fused by Dempster's rule to get modality-level mass function; (4) inside each modal, the calculated modality-level mass functions are fused by Dempster's rule; (5) decision-making is made based on the final fused mass functions and outputs a segmented mask. Here we show the segmentation example of the same located pixels $z_i^1$ and $z_i^M$ that are obtained from modal $1$ and $M$. The same pixel from different modalities is transferred separately into $H$ classifiers, and some BBAs are used to get pixel-level mass functions. Similar to Figure~\ref{fig:fig6}, we assume we can obtain $K$ mass functions with one classifier for each pixel. After the fusion of feature-level, classifier-level, and modality-level evidence fusion, a final mass function is assigned to the pixel $z_i$ to represent the degree of belief this pixel belongs to class a, b and ignorance.\label{fig:fig9}}
\end{figure*}

\begin{table}
  \centering
  \caption{Summary of BFT-based medical image segmentation methods with multimodal inputs and several classifiers/clusterers }
  \scalebox{0.7}{
  \begin{tabular}{llllllll}
 \hline
 \multicolumn{1}{l}{Publications} &Input image type& Application& BBA method &\\

\hline
Gautier et al. \cite{gautier2000belief} &multimodal MR images & lumbar sprain segmentation & Prior knowledge \\
Lajili et al. \cite{lajili2018two}&CT& breast segmentation&Threshold  \\

Huang et al. \cite{huang2021deep}&PET/CT &lymphoma segmentation &None\\
Huang et al. \cite{huang2022contextual}&multimodal MR images  &brain tumor segmentation& ENN\\

\hline
\end{tabular}
}
\label{tab:fusion4}
\end{table}
Figure \ref{fig:fig9} shows an example of a medical image segmentation framework with several classifiers and multimodal inputs, which is  more complex than the frameworks  introduced in Sections \ref{subsubsec: single}, \ref{subsubsec: multi} and \ref{subsubsec:single-modality}. The segmentation process comprises five steps: mass calculation, feature-level evidence fusion (optional), classifier-level evidence fusion, modality-level evidence fusion, and decision-making. Pixels at the same position from different modalities are fed into different classifiers and different BBA methods to get pixel-level mass functions. Dempster's rule is used first for feature-level evidence fusion, then to fuse classifier-level evidence, and last to fuse modality-level evidence. Table \ref{tab:fusion4} summarizes the segmentation methods with multimodal inputs and several classifiers or clusterers, focusing on classifier or clusterer fusion and modality-level evidence fusion.

In~\cite{gautier2000belief}, Gautier et al. proposed a method for helping physicians monitor  spinal column diseases with multimodal MR images. At first, an initial segmentation was applied with active contour~\cite{lai1994deformable}. Then several mass functions were obtained from expert opinions on constructing the frame of discernment. Thus, the mass functions were human-defined. Finally, Dempster's rule was then used to fuse the mass functions from different experts. The method yielded the most reliable segmentation maps when the paper was published.

Based on their previous work on multimodal medical image fusion with a single cluster~\cite{chaabane2009relevance}, Chaabane et al. presented another BFT-based segmentation method with several clusterers~\cite{chaabane2011new}. First, possibilistic C-means clustering~\cite{bezdek2013pattern} was used on R, G, and B channels to get three MVs. Then the MVs were mapped into mass functions with focal sets of cardinality 1 and 2 using Zhu's model~\cite{zhu2002automatic}. Dempster's rule was used first to fuse three mass functions from three corresponding color spaces. Based on the initial segmentation results, another mass function was calculated for each pixel and its neighboring pixels for each color space. Finally, the authors used Dempster's rule again to fuse the two mass functions from two corresponding clusterers. Experimental segmentation performance with cell images showed the effectiveness of the proposed method. Compared with the
FCM-based segmentation method, the proposal increased by 15\% the segmentation sensitivity.

In~\cite{lajili2018two}, Lajili et al. proposed a two-step evidential fusion approach for breast region segmentation. The first evidential segmentation results were obtained by a gray-scale-based K-means clustering method, resulting in $k$ classes. A sigmoid function was then used to define a mass function on the frame $\Omega=\{\text{Breast},\text{Background}\}$ at each pixel $z$ depending on its class.  
For local-homogeneity-based segmentation, the authors modeled the uncertainty with a threshold value $\alpha$, by defining $m(\{\text{Breast}\})=1-\alpha$, $m(\{\text{Background}\})=0$, $m(\Omega)=\alpha$, where $\alpha$ represents the belief mass corresponding to the uncertainty on the membership of the pixel $z$. A final fusion strategy with Dempster's rule was applied to combine evidence from the two mass functions. Experiments were conducted on two breast datasets~\cite{suckling1994mammographic,bowyer1996digital}. The proposed segmentation approach yielded more accurate results than the best-performed method. It extracted the breast region with correctness equal to 0.97, which was 9\% higher than the best-performing method. 

In~\cite{huang2021deep}, Huang et al. proposed to consider PET and CT as two modalities and used to UNet model to segment lymphoma separately. Then the two segmentation masks were fused by Dempster's rule. Although this is the first work that applied BFT in multimodal evidence fusion with several deep segmentation models, a limitation of this work is that only Bayesian mass functions are used for evidence fusion. To improve this first
work, Huang et al. proposed a multimodal evidence fusion framework with contextual discounting for brain tumor segmentation~\cite{huang2022contextual}. In this work, using four modules for feature extraction and evidential segmentation, the framework assigns each voxel a mass function. In addition, a contextual discounting layer is designed to take into account the reliability of each source when classifying different classes. Finally, Dempster's rule is used to combine the discounted evidence to obtain a final segmentation. This method can be used together with any state-of-the-art segmentation module to improve the final performance.

Few studies have considered multimodal medical images as independent inputs and used independent classifiers to generate mass functions. The performance of this kind of approach is limited by the representation of image features and the ability to quantify the model uncertainty. Huang et al.~\cite{huang2022contextual} first merge ENN with UNet for the fusion of multimodal MR images with contextual discounting. It enables the model to generate a learned reliability metric from input modalities during different
segmentation tasks, which can potentially make the results more explainable.

\subsection{Discussion}

The choice of using single modal or multimodal depends on the dataset. Generally, the more source data we have, the more reliable segmentation results we will get. The choice of a single or several classifiers/clusterers depends on the limitation of the computation source and the requirement of computation efficiency. Considering the main advantages of using multiple classifiers for multimodal medical images, as shown in the work of~\cite{huang2022contextual}, a potential research direction based on this approach could be promising. 

Prior to 2020, BFT-based medical image segmentation methods limited segmentation accuracy due to the use of low-level image features, such as grayscale and shape features, to generate mass functions. Moreover, none of them considered the segmentation reliability. Since the application of deep learning in medical image segmentation has been very successful, the use of BFT in deep neural networks should be a promising research direction, in particular, to quantify the uncertainty and reliability of the segmentation results.

\section{Conclusion and future work}

\label{sec:conclusion}
\subsection{Conclusion}

The main challenges in medical image segmentation are: (1) the limited resolution of medical images, (2) noisy labeling, and (3) unreliable segmentation models due to imperfect data and annotations. This leads to epistemic segmentation uncertainty and has resulted in a gap between experimental performance and clinical application for a while. For example, traditional medical image segmentation methods are limited by the use of low-level image features for decision making~\cite{kimmel2003fast,salvador2004determining,onoma2014segmentation},
leading to poor segmentation accuracy. Machine learning techniques, especially deep learning such as UNet~\cite{ronnebergerconvolutional} and its variants~\cite{trebing2021smaat,cao2021swin,hatamizadeh2022unetr}, have contributed to increasing segmentation accuracy thanks to their  ability to extract high-level semantic features. However, the clinical applicability of a deep segmentation model depends not on its segmentation accuracy but also its reliability, which is critical in the medical domain.

The issue of uncertainty and reliability of machine learning methods has recently come to the forefront~\cite{hullermeier2021aleatoric}, and has incited researchers in the medical image segmentation domain to study both accurate and reliable segmentation methods~\cite{kwon2020uncertainty,mehrtash2020confidence,ghoshal2021estimating,zhou2022trustworthy}. This review has provided a comprehensive overview of  BFT-based medical image segmentation methods. We first showed how uncertainty can be modeled in the BFT framework using mass functions, and how unreliable or conflicting sources of information can be combined to reach reliable fusion results. The segmentation methods reviewed cover a wide range of  human tissues (i.e.,~heart, lung) and tumors (i.e.,~brain, breast, and skin). In particular, we first briefly introduced medical image segmentation methods in Section \ref{sec:medical_image_segmentation} by describing the traditional and deep learning medical segmentation methods and by discussing their shortcomings and limitations. We then summarized the main concepts of BFT in Section \ref{sec:BFT}. Section \ref{sec:mass_finction} comprehensively introduced the BBA methods used to represent uncertainty (generate mass functions) in medical image segmentation tasks.  Finally, Section \ref{sec:BFT-based fusion} introduced the BFT-based medical image segmentation methods in detail by grouping them according to the number of input modalities and classifiers used to generate mass functions. 

In many applications, BFT-based methods have the potential to outperform   probability-based methods by modeling information more effectively  and combining multiple evidence at different stages, namely, at feature, modality, and classifier levels. We hope that this review can increase awareness and understanding of BFT theory and how it can  contribute to the development of more accurate and reliable medical image segmentation techniques.

\subsection{Future work}

Despite the advantages of BFT for medical image segmentation, existing methods still have limitations that need to be addressed. Some directions for further research are discussed below.

First, most of the existing BFT-based medical image segmentation methods still use low-level features and do not fully exploit the advantages of deep learning. Combining BFT with deep segmentation networks should allow us to develop both accurate and reliable segmentation models, particularly in medical image segmentation tasks for which medical knowledge is available and can be modeled  by mass functions. Ref.~\cite{george2020breast} is the first successful application of Dempster's rule for the fusion of different CNNs. Ref.~\cite{huang2022lymphoma} confirmed the similarity of ENN and RBF when acting as an evidential classifier and integrated both classifiers within a deep segmentation network to improve the segmentation accuracy and reliability. We believe that more promising results will be obtained by blending BFT with the existing powerful deep medical image segmentation models~\cite{trebing2021smaat,cao2021swin,hatamizadeh2022unetr}.

Second, even though the BFT-based fusion methods have considered the conflict of multiple sources of evidence,  a problem still remains in that  multimodal medical images may have different degrees of reliability when segmenting different regions. For example, for the BRATS brain tumor dataset\footnote{\url{http://braintumorsegmentation.org/}}, 
domain knowledge from the physicians tells us that areas corresponding to the enhancing tumor are described by faint and enhancement on T1Gd MR images, and edema is recognized by the abnormal hyperintense signal envelope on the FLAIR modality~\cite{baid2021rsna}. Multimodal medical image fusion  may fail if some source of information is unreliable or partially reliable for performing different segmentation tasks. In~\cite{huang2022contextual}, Huang et al. proposed a new method to learn reliability coefficients conditionally on different segmentation tasks in a deep neural network. Another interesting research direction, accordingly, is to estimate task-specific reliability and to enhance the explainability of deep evidential neural networks using contextual discounting for multimodal or cross-modal medical image segmentation tasks.

Third, acquiring large amounts of labeled training data is particularly challenging for medical image segmentation tasks and has become the bottleneck of learning-based segmentation performance. The successful application of  unsupervised BBA methods to medical image segmentation points to a new direction to address the lack of annotated data. As far as we know, there is no published paper dealing with the combination of unsupervised BBA methods with deep learning. The  neural-network-based evidential clustering method described in~\cite{denoeux21b} and the EKNN rule with partially supervised learning introduced in~\cite{denoeux2019new} are two steps in these directions. These works provide insights into how to use unsupervised or semi-supervised learning to quantify segmentation uncertainty with unannotated or partially annotated data sets.

We hope this review will increase awareness of the challenges of existing BFT-based medical image segmentation methods and call for future contributions to bridge the gap between experimental performance and clinical application, as well as to develop accurate, reliable, and explainable deep segmentation models. 
\section*{Acknowledgements} 
The first author received funding from the China Scholarship Council (No.201808331005). This work was carried out in the framework of the Labex MS2T, which was funded by the French Government through the program ``Investments for the future'' managed by the National Agency for Research (Reference ANR-11-IDEX-0004-02)

\bibliographystyle{elsarticle-num} 
\bibliography{references}

\end{document}